%% file: labvla.tex
\documentclass[]{labvla}
\usepackage{fix-cm}
\usepackage{helvet}

\usepackage{xcolor}
\usepackage{multirow}
\usepackage{colortbl}
\usepackage{tcolorbox}
\usepackage{enumitem}
\usepackage{pifont}
\usepackage{wrapfig}
\usepackage{url}

\definecolor{mygray}{gray}{0.9}
\definecolor{myblue}{HTML}{F0FFFF}
\definecolor{myblue_light}{HTML}{EAF6FF}
\definecolor{myblue_dark}{HTML}{2B57A0}
\definecolor{myblue_lightv1}{HTML}{F2F7E0}
\definecolor{mygreen}{RGB}{60, 179, 113}
\definecolor{myyellow_light}{RGB}{255, 255, 224}
\definecolor{lightgreen}{HTML}{D3E5C7}
\definecolor{deepgreen}{RGB}{46,139,87}

\newcommand{\hlalt}[1]{\textbf{\textcolor{accent!80!black}{#1}}}

\newcommand\blfootnote[1]{%
  \begingroup
  \renewcommand\thefootnote{}\footnote{#1}%
  \addtocounter{footnote}{-1}%
  \endgroup
}

\title{LabVLA: Grounding Vision-Language-Action Models in Scientific Laboratories}

\author{
    \mbox{Baochang Ren\textsuperscript{1}},
    \mbox{Xinjie Liu\textsuperscript{1}},
    \mbox{Xi Chen\textsuperscript{1}},
    \mbox{Yanshuo Liu\textsuperscript{1}},
    \mbox{Chenxi Li\textsuperscript{2}},
    \mbox{Daqi Gao\textsuperscript{1}},
    \mbox{Zeqin Su\textsuperscript{1}},
    \mbox{Jintao Xing\textsuperscript{1}},
    \mbox{Zirui Xue\textsuperscript{1}},
    \mbox{Rui Li\textsuperscript{2}},
    \mbox{Xiangyu Zhao\textsuperscript{1}},
    \mbox{Shuofei Qiao\textsuperscript{1}$\dagger$},
    \mbox{Minting Pan\textsuperscript{2}},
    \mbox{Wangmeng Zuo\textsuperscript{3}},
    \mbox{Lei Bai\textsuperscript{2}},
    \mbox{Dongzhan Zhou\textsuperscript{2}$\dagger$},
    \mbox{Ningyu Zhang\textsuperscript{1}$\dagger$},
    \mbox{Huajun Chen\textsuperscript{1}}
}
\affiliation[1]{\mbox{Zhejiang University}}
\affiliation[2]{\mbox{Shanghai AI Laboratory}}
\affiliation[3]{\mbox{Harbin Institute of Technology}}

\abstract{
Scientific laboratories increasingly rely on AI systems to reason about experiments, but the physical act of doing science remains largely outside their reach. AI can help read literature, generate hypotheses, and plan protocols, yet the execution of those protocols at the bench still requires a human operator. Vision-Language-Action (VLA) models provide one possible interface between written protocols and robot execution, but existing policies are trained mostly on household and tabletop demonstrations and rarely encounter the instruments, transparent liquids, or fixed protocol workflows found in scientific laboratories. Closing this gap requires both laboratory-specific supervision and a unified learning framework that can accommodate the diverse robot embodiments used to execute experimental protocols. We therefore identify data and embodiment as central bottlenecks alongside model design. To address the data side, we build RoboGenesis, a simulation-based workflow and data engine that composes configured laboratory workflows from atomic skills, validates and filters rollouts, and exports structured demonstrations across supported robot profiles. On the policy side, we present LabVLA, trained with a two-stage recipe: FAST action token pretraining first makes the Qwen3-VL-4B-Instruct backbone action aware before any continuous control is learned, and flow matching posttraining then attaches a DiT action expert under knowledge insulation. On the LabUtopia benchmark, LabVLA achieves the highest average success rate among all evaluated baselines under both in-distribution and out-of-distribution settings.
}

\teaser{%
    \includegraphics[width=0.98\linewidth]{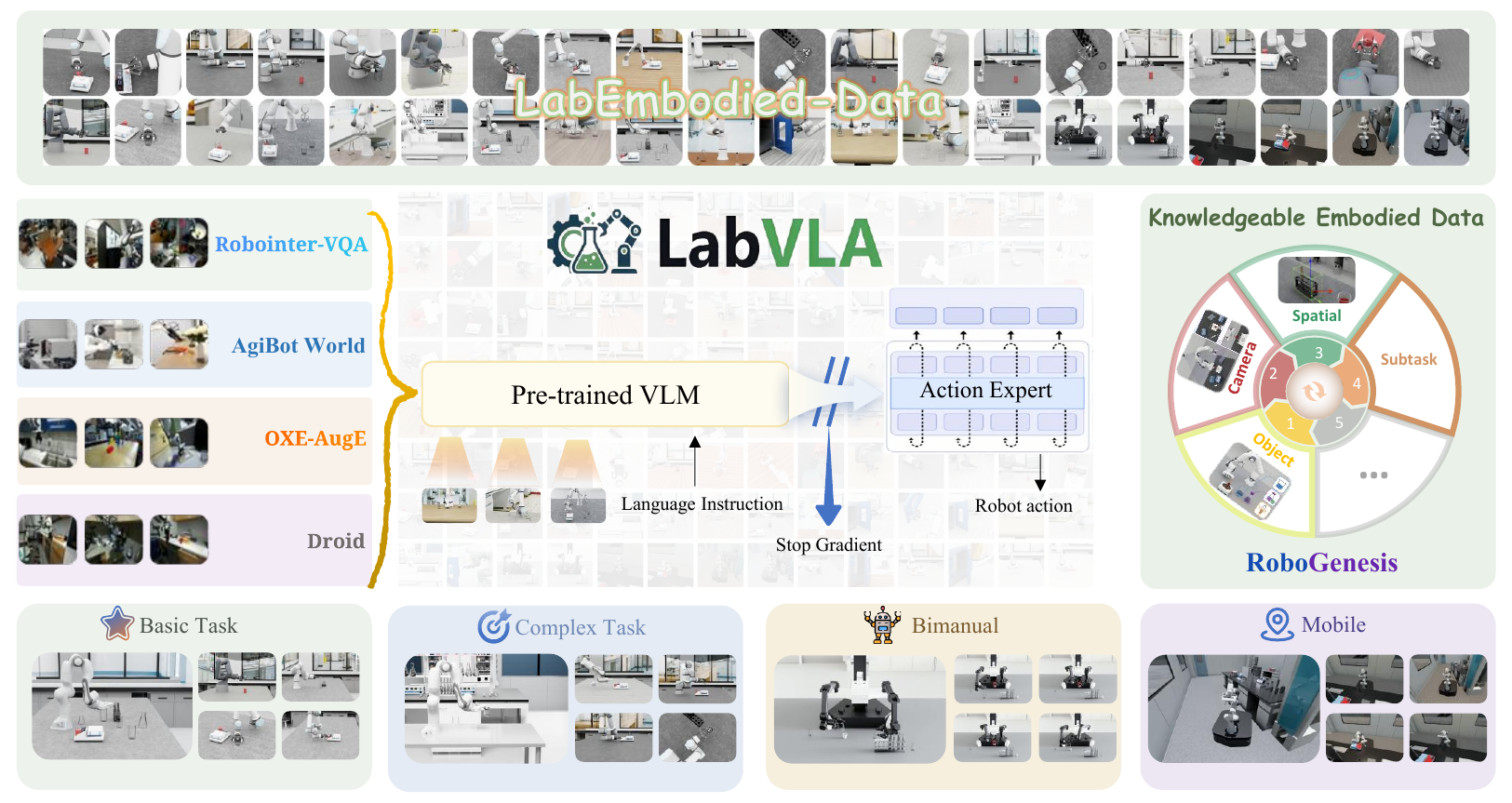}%
    \captionsetup{hypcap=false,font=footnotesize}%
    \captionof{figure}{\textbf{LabVLA framework at a glance.}
    \textbf{Center:} the LabVLA policy couples a pretrained VLM, which encodes language instructions together with multiview images, to a DiT action expert that emits robot actions; a stop-gradient (knowledge insulation) reduces interference from the action loss while the action expert specializes.
    \textbf{Top:} representative scenes from LabEmbodied-Data, the supported embodiment laboratory demonstration corpus synthesized by RoboGenesis.
    \textbf{Left:} external corpora that warmstart the visual-language backbone (Robointer-VQA, AgiBot World Beta, OXE-AugE, and Droid).
    \textbf{Right:} RoboGenesis, our simulation based workflow and data engine, operates in three stages: (1)~environment building, which populates a 3D asset library via text to image generation and TRELLIS~2.0 reconstruction, then assembles validated laboratory scenes; (2)~agentic workflow generation, which decomposes a natural language instruction into an ordered sequence of atomic skills and instantiates it across 10+ robot platforms with six axis domain randomization; and (3)~structured export, which filters rollouts by execution success and attaches diverse annotation streams.
    \textbf{Bottom:} LabEmbodied-Data covers four task families: single arm primitives, multistep laboratory procedures, bimanual operations, and mobile manipulator settings.}%
    \label{fig:overview}%
}

\badge{\faHome}{Homepage}{https://zjunlp.github.io/LabVLA}
\badge{\faGithub}{Code}{https://github.com/zjunlp/LabVLA}
\badge{\faHuggingFace}{Model}{https://huggingface.co/zjunlp/LabVLA}
\badge{\faEnvelope}{Contact}{mailto:zhangningyu@zju.edu.cn}

\begin{document}

\blfootnote{\vspace{4mm}$^\dagger$Corresponding author.}
\maketitle

\input{section/1.intro}

\input{section/data}

\input{section/method}

\input{section/experiment}

\input{section/analysis}

\input{section/related_work}

\input{section/conclusion}

\input{section/discussion}

\bibliographystyle{plainnat}
\bibliography{reference}

\appendix
\input{section/appendix}

\end{document}

%% file: section/1.intro.tex
\newpage

\section{Introduction}

AI for research is beginning to change how scientists search literature, write code, design hypotheses, and plan experiments~\cite{jumper2021highly,taylor2022galactica,merchant2023scaling,lu2024ai}.
Self-driving laboratory research~\cite{tom2024self} and language model laboratory agents~\cite{boiko2023autonomous,m2024augmenting} have begun to extend this assistance from digital reasoning to physical experimentation.
The physical phase of an experiment, in which a robot picks up a beaker, transfers a reagent, presses a heater button, and watches for a color change, still falls almost entirely to a human operator at the bench.
The gap between digital scientific reasoning and real experimental work is therefore not one of intent but of embodiment.
Vision-Language-Action (VLA) models provide a possible interface for this gap, but they have been trained mostly on household and tabletop demonstrations, which leaves them without the instrument knowledge, contact precision, and protocol level supervision that benchtop procedures such as reagent preparation or genetic analysis require.

Laboratory manipulation differs from ordinary tabletop manipulation in its failure modes.
Pipette aspiration, cap screwing, liquid transfer, and button operation demand fine spatial precision and reliable contact control.
Many tasks depend on physical state changes such as liquid flow, heating, mixing, color transitions, and container placement.
The same protocol must also run on different robot embodiments with different cameras, end effectors, workspaces, and action dimensions.
These factors make data and embodiment central bottlenecks in laboratory automation, alongside the design of the policy itself.

Existing robot corpora give VLA training a broad manipulation prior.
Open X-Embodiment~\cite{o2024open}, DROID~\cite{khazatsky2024droid}, and BridgeData V2~\cite{walke2023bridgedata} have driven progress in household and tabletop manipulation and supply VLA policies with general robot operation priors over objects, viewpoints, embodiments, and contact patterns.
Open policies such as OpenVLA~\cite{kim2024openvla} and $\pi_0$~\cite{black2024pi_0} have successfully scaled on these corpora.
However, these corpora rarely include pipettes, centrifuges, thermal cyclers, heating plates, transparent liquids, or protocol level chemistry and biology workflows, so a policy trained only on them cannot ground a written laboratory protocol in the right instruments and physical states.
Collecting such data directly in real laboratories is expensive: it requires specialized instruments, domain supervision, calibrated hardware, and strict safety procedures, which together push the data collection cost far above that of ordinary robot data collection.

To make this setting trainable, we build \textbf{RoboGenesis}, an Isaac Sim based data synthesis engine for laboratory automation.
Training a VLA primarily from simulation is a deliberate choice.
Simulation side data collection scales with compute rather than with instrument, supervision, and safety overhead, so each configured protocol can be replayed across randomized scenes and supported robot profiles at a marginal cost far below the real laboratory data collection noted above.
InternData-A1~\cite{tian2025interndata} recently showed that a policy trained almost entirely on synthetic demonstrations can match $\pi_0$ on real robot evaluations, which suggests that high fidelity simulation data can substitute for real laboratory collection when scenes, physics, and protocols are faithfully modeled.
Inspired by this, RoboGenesis first constructs executable laboratory scenes from instruments, containers, consumables, and robot assets, then composes protocol workflows from atomic manipulation skills, randomizes visual and spatial factors, and filters demonstrations by execution success.
Unlike fixed simulation corpora, RoboGenesis lets users programmatically specify configured workflows for laboratory protocols and instantiate them on supported single-arm, mobile-manipulator, or dual-arm robot profiles, with domain randomization layered on top and success filtering applied before export.
We use this engine to synthesize LabEmbodied-Data, a corpus of multi camera observations, language instructions, robot states, action trajectories, and structured annotations under a shared cross-embodiment schema.

Building on this corpus together with broad real-robot pretraining data, we present \textbf{LabVLA}, a VLA pipeline for connecting written laboratory protocols to embodied robot execution in simulated scientific workspaces.
LabVLA pairs protocol conditioned data synthesis with FAST~\cite{pertsch2025fast} action token pretraining and flow matching~\cite{lipman2022flow} posttraining under a shared cross-embodiment schema.
\Cref{fig:overview} summarizes the end-to-end LabVLA pipeline: broad visual-language and manipulation priors are first acquired from web-scale and real-robot corpora, then adapted to laboratory protocol execution using structured LabEmbodied-Data synthesized by RoboGenesis, and finally evaluated across four LabUtopia~\cite{li2026labutopia} task families.
Specifically, LabVLA adapts a Qwen3-VL~\cite{bai2025qwen3} backbone to map visual observations, robot state, and language instructions into continuous action chunks through a DiT~\cite{peebles2023scalable} action expert.
The model is trained in two stages: FAST action tokens first align the visual language prefix with action semantics during VLM pretraining, and flow matching then predicts continuous robot actions during posttraining.
A knowledge insulation design (the stop-gradient~\cite{driess2026knowledge} arrow in \Cref{fig:overview}) reduces interference between language grounded VLM representations and the continuous action expert during posttraining.

In summary, our contributions are:
\begin{itemize}[leftmargin=*,itemsep=1pt,topsep=2pt]
    \item We formulate scientific laboratory automation as a VLA learning problem and argue that data and embodiment are central bottlenecks alongside model design.
    \item We introduce RoboGenesis, a simulation based workflow and data engine that links environment construction, configured workflow generation, domain randomization, and success filtered export to produce laboratory demonstrations that existing robot corpora rarely cover.
    \item We present LabVLA, a Qwen3-VL based policy that combines FAST action token pretraining, flow matching posttraining, and knowledge insulation for fixed laboratory protocol execution.
\end{itemize}

%% file: section/data.tex
\section{RoboGenesis: A Programmable Workflow and Data Engine}\label{sec:robogenesis}

\begin{figure*}[t]
    \centering
    \includegraphics[width=0.98\linewidth]{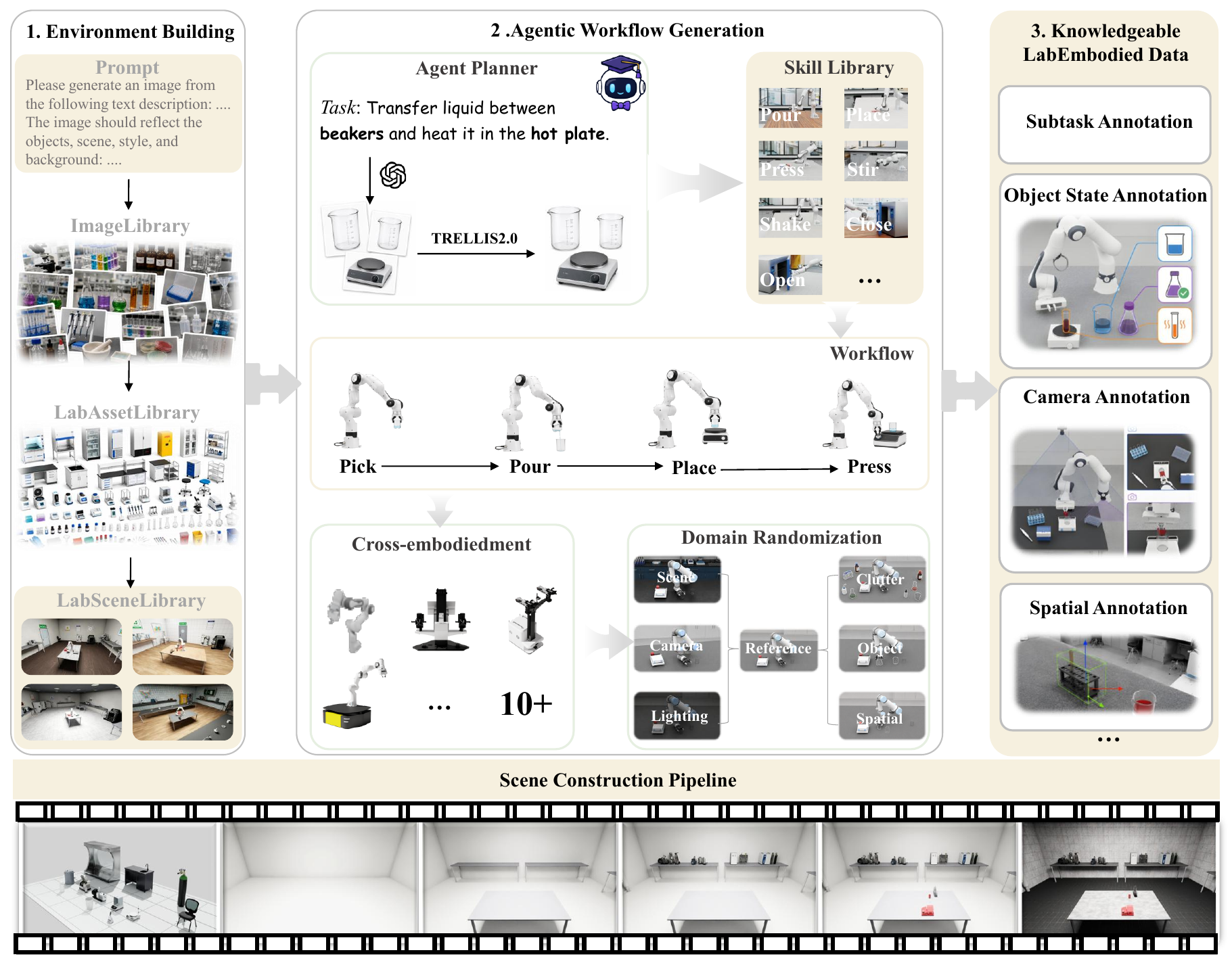}
    \caption{\textbf{RoboGenesis data synthesis pipeline.}
    \textbf{(1)~Environment Building} (left): text prompts produce reference images that pass through TRELLIS~2.0 to populate LabAssetLibrary; a scene construction pipeline assembles validated scenes (LabSceneLibrary, bottom strip).
    \textbf{(2)~Agentic Workflow Generation} (center): an agent planner decomposes a natural language instruction into an ordered workflow of atomic skills, e.g.\ \texttt{Pick}(beaker1) $\rightarrow$ \texttt{Pour}(beaker1, beaker2) $\rightarrow$ \texttt{Place}(hot\_plate) $\rightarrow$ \texttt{Press}; the workflow is instantiated across 16 robot platforms (cross embodiment) and diversified through domain randomization over scene, camera, lighting, clutter, object, and spatial factors.
    \textbf{(3)~Knowledgeable LabEmbodied-Data} (right): successful rollouts are exported with subtask, object state, camera, spatial, and additional annotations.}
    \label{fig:robogenesis}
\end{figure*}

Building a laboratory grade VLA pipeline requires training data with three properties at once: executable laboratory scenes, reusable protocol structure, and trajectories paired with the task information that protocol aware policies need.
Existing robot corpora do not provide this combination, and direct real laboratory collection is difficult to scale because it depends on specialized instruments, calibrated hardware, safety procedures, and domain supervision.
We address this data bottleneck with RoboGenesis, a simulation based workflow and data engine for scripted robot manipulation.
The architecture itself is domain general; we deploy it here for laboratory tasks.
As shown in \Cref{fig:robogenesis}, RoboGenesis has three stages: environment building, agentic workflow generation with cross embodiment deployment and domain randomization, and structured export into knowledgeable LabEmbodied-Data.
These stages turn a reviewed workflow into success filtered demonstrations over supported instruments, objects, layouts, cameras, and robot profiles.
To situate RoboGenesis among recent synthetic data and simulation engines, \Cref{tab:robogenesis_comparison} compares design mechanisms that matter for protocol conditioned data generation rather than broad policy performance.
The comparison covers workflow executability, long horizon task composition from atomic skills, success filtered export, and annotation granularity.

\input{tables/robogenesis}

Among the engines in \Cref{tab:robogenesis_comparison}, RoboTwin~2.0~\cite{chen2025robotwin} automates task code generation with domain randomization across five embodiments, but each task is a monolithic script with fixed scene layouts, and its exported annotations stop at subtask indices.
RoboCasa~365~\cite{nasiriany2026robocasa365} ships AI generated assets and LLM proposed tasks, yet both are produced offline: the released engine samples from a library of 2{,}500+ prebuilt scenes and 365 hardcoded task classes, and its demonstrations cover a single PandaOmron embodiment.
ManiSkill~3~\cite{tao2024maniskill3} and RLBench~\cite{james2020rlbench} are mature multi embodiment simulators (23 and 5 robot platforms) with built in domain randomization and per task success checking, but neither generates assets, scenes, or tasks, and all tasks are self contained classes with no composition mechanism.
RoboGen~\cite{wang2023robogen} proposes scenes and tasks with a language model, yet retrieves assets from fixed libraries, randomizes only object placement, and binds each generated task to one robot with hardcoded substep sequences.
RoboGenesis is the only entry that checks every column: generative asset creation, agent based scene and task generation, configurable domain randomization, per skill success filtering, structured annotations, long horizon composition from an atomic skill library, and lab protocol support across 16 robot platforms.
No other engine allows users to freely chain atomic skills into new long horizon workflows; together with end to end asset generation, this composability is the feature most specific to our use case.
The same workflow definitions also serve as evaluation protocols, so users can benchmark a policy on any task they compose without additional code.

\subsection{Environment Building}
Every trainable demonstration must begin from a physically executable environment.
If the base scene has unreachable placements, missing instruments, unstable contacts, or invalid object geometry, downstream workflow composition and randomization only produce more unusable rollouts.
RoboGenesis builds each environment in two stages: first it populates an asset library of simulation ready 3D objects, then it assembles those assets into validated laboratory scenes.

\paragraph{Asset generation pipeline.}
Building an asset library by hand is impractical at the scale and category range a laboratory demands.
RoboGenesis instead converts a plain text description into a physics annotated USD asset through four steps:
(1)~Given a text description of the target object, the pipeline constructs a structured product-photography prompt that specifies material, viewpoint, lighting, and background, then calls a text to image API to render the reference photograph (the full prompt template is listed in \Cref{app:asset-prompt}).
(2)~The reference image is fed to TRELLIS~2.0~\cite{xiang2026native}, a feedforward image to 3D model that reconstructs a textured mesh in GLB format.
(3)~A postprocessing stage orients the mesh upright, exports it to USD with PBR textures, optionally decimates the triangle count with MeshAnythingV2, and generates a collision mesh together with a URDF carrying mass, friction, and bounding box metadata drawn from a size catalog.
(4)~The pipeline runs in batch mode across diverse equipment categories and produced the LabAssetLibrary, a pool of 2{,}947 annotated assets that RoboGenesis draws from when composing scenes.
Containers that require visible liquid declare a color and fill fraction in the workflow configuration; the engine generates a colored mesh proxy inside each vessel and transfers the liquid from source to target when a pour step succeeds.

\paragraph{Automated scene construction pipeline.}
Given the asset library, RoboGenesis assembles laboratory scenes through a seed driven, greedy placement pipeline whose numerical constraints are listed in \Cref{app:placement-rules}.
The pipeline first scans the asset catalog and measures each mesh's bounding box against a category aware size prior, classifying every asset as \emph{table} ($\leq$\,0.4\,m, e.g.\ beakers), \emph{bench} (midsize instruments), or \emph{floor} ($\geq$\,0.5\,m, e.g.\ cabinets).
A seeded random generator then produces a scene intent: room dimensions, a central table topology (perimeter, island, or parallel row), four functional wall themes drawn from a pool (wet chemistry, thermal, measuring, bio, storage), and a door wall.

The solver places assets sequentially in six passes, checking each candidate against all previously placed objects via axis aligned bounding box overlap and gap tests; any candidate that violates a constraint is skipped.
(1)~The main task table is pinned at the origin beneath the robot so that all task objects are reachable; 1--3 extra work tables are added according to the chosen topology, each enforcing a minimum robot aisle gap to existing tables.
(2)~The solver walks each wall from corner to corner, alternating between lab counters and floor standing furniture while skipping the door keepout zone and corner margins.
(3)~Each counter is populated with themed clusters of 2--4 bench class assets sharing a functional category (e.g.\ a weighing station or a titration bench), with tight intracluster spacing and wider intercluster gaps.
(4)~Floor standing equipment (cabinets, fume hoods) is placed against walls; each item is collision checked against counters, tables, doorways, and other floor items, with a guaranteed robot aisle between perimeter furniture and central tables.
(5)~Shelves and signage fill the upper wall above each counter, and small glassware is scattered on the extra work tables.
(6)~All wall adjacent assets receive a canonical yaw computed from their front axis metadata so they face into the room.

After placement, the solver runs ten validation checks (work table clearance, robot aisle width, robot placement on the main table, counter presence, floor asset size and count, surface grounding for both table and floor items, overlap free placement, and wall clip avoidance) and computes a 0--100 quality score; scenes below threshold are rejected and re-seeded.
Physics annotation then attaches rigid body dynamics, convex decomposition collision meshes, mass, and friction to every interactive object, while static surfaces receive invisible box colliders for stable contact.
The output is a validated USD scene that the randomization stage can diversify per episode without revisiting the layout.
We also curated over 1{,}000 texture images as the LabTextureLibrary for surface material assignment.
Using the LabAssetLibrary and LabTextureLibrary, we generated 10{,}000 laboratory scenes; the pipeline itself imposes no such ceiling; given the combinatorial space of assets, textures, topologies, and wall themes, a single batch run could produce 100{,}000 scenes or more.
We chose 10{,}000 as a practical trade-off between interscene diversity and storage cost; users who expand the asset and texture libraries can generate proportionally more.
Representative examples of the resulting scene diversity are shown in \Cref{app:scene-diversity}.

\paragraph{Robot profiles.}
The robot profile pool lives separately from the scene and protocol definitions.
It covers single arm, bimanual, and mobile manipulator settings, including Franka Panda, FR3, UR-series arms, Piper, Rizon4, Festo, ARX X5, ARX R5, Split ALOHA, Lift2, FR3 Duo, and Ridgebase-mounted variants.
Each configuration profile stores the robot's kinematic parameters, gripper settings, camera frames, and skill level overrides separately from the protocol.
This separation lets the same reviewed workflow run on different embodiments through robot specific overrides; the rollout is accepted only when it passes its success checks.

\paragraph{Interactive authoring.}
Beyond batch generation, RoboGenesis supports interactive authoring of both assets and scenes through a web based Task Designer.
For assets, a user can supply a single reference image and the pipeline will extract a 3D mesh via TRELLIS~2.0, apply postprocessing, and register the result into the asset library, which is useful for adding objects not covered by the batch categories.
For scenes, the Task Designer offers a scene builder for selecting assets and furniture from the catalog and applying textures, a manual configuration mode for direct control of object positions, robot placement, and camera setup, and a drag and drop mode for visual repositioning.
All modes share a unified asset registry and export validated scene configurations that the downstream pipeline can consume directly.

\subsection{Agentic Workflow Generation}
Laboratory protocols are long horizon sequences of atomic manipulations.
A robot may need to pick up a beaker, pour into a target container, return the source, move the reaction vessel to a heater, press a button, pick up a glass rod, stir, return the rod, and then remove or shake the final product.
Hardcoding such a procedure for each scene and robot would not scale.
RoboGenesis instead represents a protocol as a workflow template containing a natural language instruction, named scene objects, target references, and an ordered list of atomic skills.

\paragraph{Atomic skill library.}
The skill library covers object manipulation, instrument interaction, and navigation.
It includes object centric skills such as \texttt{pick}, \texttt{place}, \texttt{pour}, \texttt{stir}, \texttt{shake}, and \texttt{move}, instrument level skills such as \texttt{press}, \texttt{pressZ}, \texttt{open}, and \texttt{close}, and navigation skills for mobile embodiments.
The skills operate at the laboratory action level rather than at the controller level, so a reviewed workflow can be reused while different robot profiles adapt low level execution.
The library is not closed: users can register additional atomic skills to extend the repertoire beyond the built-in set.

\paragraph{Workflow authoring.}
RoboGenesis provides an agent assisted authoring path that turns a natural language instruction into an executable workflow.
Given an instruction such as ``transfer liquid between beakers and heat it in the lab setup'', the agent queries the available robot, task specification, and catalog libraries, selects a matching template, generates scene layouts, and produces a candidate workflow YAML.
An offline validator then scores the candidate by checking pick zone reachability, crowded place target conflicts, and repick yaw risks; if the score falls below a threshold, the agent is allowed one retry with different template or layout choices before the workflow is returned for human review.
Users can also define workflow YAML templates by hand, specifying scene objects (names, USD paths, position ranges, and semantic roles), an ordered list of skill steps with per step target objects and parameters, and a natural language instruction for the overall task.
This manual path gives full control over the protocol structure when the agent assisted route does not cover a desired procedure.
In our experiments, composite workflows exceeding 20 skill steps still achieve collection success rates above 75\%.

\paragraph{Domain randomization.}
Once a workflow has been validated, RoboGenesis instantiates it across target robot platforms and diversifies the resulting demonstrations through domain randomization.
Policies must tolerate changes in embodiment, lighting, container appearance, instrument placement, background clutter, and camera setup, but uncontrolled randomization can break the causal link between the written protocol and the executed behavior.
RoboGenesis therefore adds diversity only after both the scene and workflow pass their validation checks.
Starting from a validated reference scene, it supports six configurable randomization axes:
\textbf{scene} randomization varies the workspace layout;
\textbf{visual clutter} adds nonprotocol objects without changing the target task;
\textbf{camera} randomization perturbs the position or orientation of recording views;
\textbf{object} randomization swaps compatible assets while preserving each object's semantic role;
\textbf{lighting} randomization changes intensity, color temperature, and HDRI background;
\textbf{spatial} randomization perturbs object poses within validated placement ranges.
The engine also paraphrases the language instruction so that the policy sees surface variation in protocol descriptions.
Which axes are active is a per collection choice.

The constraint is that randomization never rewrites the experiment.
A source beaker remains the source beaker, the heat button remains attached to the heating device, and a glass rod remains associated with its rack.
Clutter is kept outside the task object contract used for labels and bounding boxes.
This preserves protocol semantics but still gives the model visual, spatial, linguistic, and embodiment variation.
Distribution shift evaluation therefore measures generalization rather than label noise introduced by the data engine.

\subsection{Knowledgeable LabEmbodied-Data}
Downstream VLA training needs demonstrations that preserve the structure of the laboratory protocol rather than raw image action pairs alone.
RoboGenesis therefore exports only successful rollouts as annotated \emph{LabEmbodied-Data}.
The engine discards failed episodes before writing training data, while episode summaries and step level pass rates remain available for debugging and analysis.
Each skill has a dedicated success checker that evaluates physical conditions specific to the operation, such as grasp stability for pick, liquid transfer for pour, or position tolerance for place.
A contact safety monitor runs alongside each checker and rejects steps that involve forbidden collisions regardless of whether the primary condition was met.

Each saved episode stores multicamera RGB observations, robot joint states, executed actions, and the language instruction.
RoboGenesis ships 15 individually configurable annotation providers:
(1)~\textbf{robot state}: per frame gripper position, open/closed ratio, joint positions, and end effector pose (position + quaternion);
(2)~\textbf{camera intrinsics}: per camera $3\times3$ intrinsics matrix, vertical field of view, and view label (wrist, top-down, third-person, side);
(3)~\textbf{camera extrinsics}: $4\times4$ extrinsic matrices with camera role assignments;
(4)~\textbf{step timing}: per step frame boundaries, skill type, target object, number of frames, and wall clock duration;
(5)~\textbf{instruction alignment}: per frame templated subinstruction derived from the active skill and target object;
(6)~\textbf{object state}: per frame state codes for each scene object (idle, picked, placed, heated, opened, closed);
(7)~\textbf{scene relations}: pairwise spatial relations between objects at episode end (left-of, right-of, above, below, inside, near);
(8)~\textbf{object semantics}: per object category, material, color, affordance list, and task relevance flag;
(9)~\textbf{success explanation}: per step success/fail verdict with checker specific metrics (lift margin, position offset, tilt angle, etc.);
(10)~\textbf{collision events}: contact start/end events between object pairs with gripper proximity distance;
(11)~\textbf{temporal segments}: segment boundaries aligned to skill transitions, with quality scores, mistake flags, and confidence;
(12)~\textbf{subgoals}: frame indices at segment boundaries marking subgoal completion points;
(13)~\textbf{quality scores}: per frame quality rating and episode level summary (mean, minimum);
(14)~\textbf{intervention flags}: per frame binary indicator of human intervention;
(15)~\textbf{episode metadata}: robot type, dataset source, control mode, action representation, collection FPS, language instruction, and overall quality score.
These annotations let the model learn both the action trajectory and its position inside the laboratory procedure.
The result is a protocol conditioned dataset with a shared observation action schema across instruments, scenes, and supported embodiments.
Because the workflow remains the unit of supervision, policy training can compare or mix compatible robot data without rewriting the scientific procedure for each robot.

%% file: tables/robogenesis.tex
\begin{table*}[t]
\centering
\caption{Feature level comparison of robot simulation and data generation engines, verified against each engine's public code release.
\ding{51} denotes features supported in the released code; -- denotes features that are not supported.}
\label{tab:robogenesis_comparison}
\vspace{-2mm}
\renewcommand{\arraystretch}{1.18}
\resizebox{0.98\textwidth}{!}{%
\begin{tabular}{l c c c c c c c c c}
\toprule
\textbf{Engine} &
\shortstack{\textbf{\# Robot}\\\textbf{/ embodiment}} &
\shortstack{\textbf{Auto}\\\textbf{asset}} &
\shortstack{\textbf{Auto}\\\textbf{scene}} &
\shortstack{\textbf{Auto}\\\textbf{task}} &
\shortstack{\textbf{Domain}\\\textbf{random.}} &
\shortstack{\textbf{Success}\\\textbf{QA}} &
\shortstack{\textbf{Structured}\\\textbf{annotations}} &
\shortstack{\textbf{Long-horizon}\\\textbf{composition}} &
\shortstack{\textbf{Lab}\\\textbf{protocol}} \\
\midrule
RoboTwin 2.0~\cite{chen2025robotwin} & 5 & -- & \ding{51} & \ding{51} & \ding{51} & \ding{51} & -- & -- & -- \\
RoboCasa 365~\cite{nasiriany2026robocasa365} & 1 & -- & \ding{51} & -- & \ding{51} & \ding{51} & -- & -- & -- \\
ManiSkill 3~\cite{tao2024maniskill3} & 23 & -- & -- & -- & \ding{51} & \ding{51} & -- & -- & -- \\
RLBench~\cite{james2020rlbench} & 5 & -- & -- & -- & \ding{51} & \ding{51} & -- & -- & -- \\
RoboGen~\cite{wang2023robogen} & 6 & -- & \ding{51} & \ding{51} & -- & \ding{51} & -- & -- & -- \\
\rowcolor{blue!6}
\textbf{RoboGenesis (Ours)} & 16 & \ding{51} & \ding{51} & \ding{51} & \ding{51} & \ding{51} & \ding{51} & \ding{51} & \ding{51} \\
\bottomrule
\end{tabular}%
}
\end{table*}

%% file: section/method.tex
\section{LabVLA Training Recipe}\label{sec:method}

\begin{figure*}[t]
    \centering
    \includegraphics[width=0.98\linewidth]{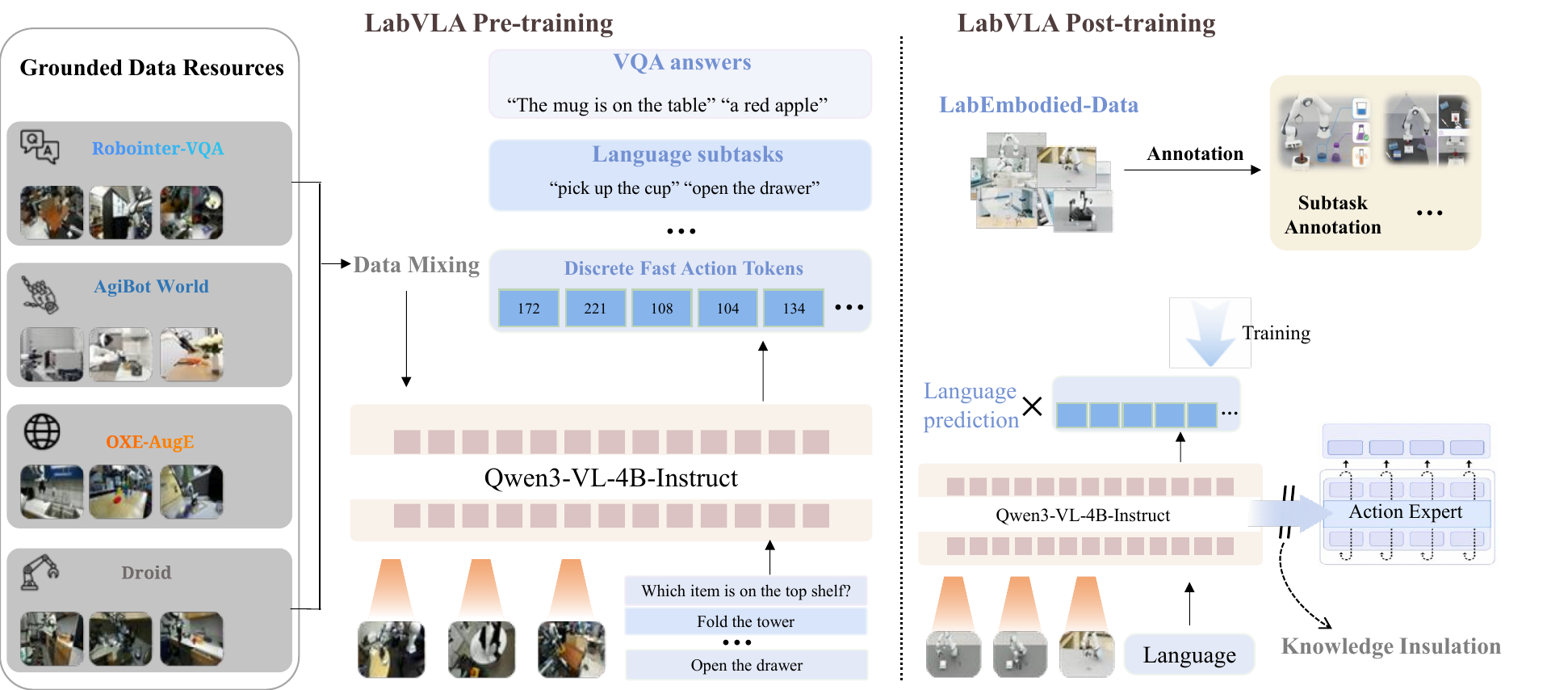}
    \caption{\textbf{LabVLA training recipe.}
    \textbf{(Left) Pretraining.} The Qwen3-VL-4B-Instruct backbone trains on grounded data sources (Robointer-VQA, AgiBot World Beta, OXE-AugE, and Droid), supervised on VQA answers, language subtasks, and discrete FAST action tokens to align the visual language prefix with action semantics before the action expert is attached.
    \textbf{(Right) Posttraining.} On OXE-AugE together with LabEmbodied-Data, the same VLM is paired with a DiT action expert that produces continuous action chunks; a stop-gradient between the VLM hidden states and the flow matching loss implements knowledge insulation to reduce interference from the action objective while the action expert specializes.}
    \label{fig:framework}
\end{figure*}

The LabVLA training pipeline runs in two stages: pretraining and posttraining, summarized in \Cref{fig:framework}.
In the \textit{pretraining} stage, a Qwen3-VL-4B-Instruct backbone is jointly trained on grounded data sources (Robointer-VQA, AgiBot World Beta, OXE-AugE, and Droid) to produce VQA answers, language subtasks, and discrete FAST action tokens, so the visual language prefix becomes action aware before any continuous action head is attached.
For OXE-AugE we use only its LeRobot format subset, which merges six of its source datasets into roughly 572k trajectories.
In the \textit{posttraining} stage, the pretrained VLM is paired with a DiT action expert and supervised on OXE-AugE together with synthesized LabEmbodied-Data built on RoboGenesis.
Flow matching predicts continuous action chunks, and a stop-gradient between the VLM hidden states and the flow matching loss implements knowledge insulation to reduce interference between token-level VLM objectives and velocity space action learning.
The remainder of this section details the architecture, the FAST pretraining loss (\Cref{sec:vlm-pretrain}), and the flow matching posttraining objective with knowledge insulation (\Cref{sec:flow-post,sec:ki}).

\paragraph{Architecture.}
As shown in \Cref{fig:framework}, LabVLA pairs a Qwen3-VL-4B-Instruct backbone with a DiT action expert.
We use $\phi$ as the VLM parameters and $\theta$ as the action expert.
For an embodiment or dataset source $r$, let $d_r$ be its active action dimension and $d_{\max}$ be the padded maximum action dimension used for batching.
At step $t$, the model observes up to $V$ RGB camera views $I_t^{1:V}$, a language instruction $\ell$, and a robot state $q_t^r$, and predicts a $K$-step continuous action chunk:
\begin{equation}
    A_t^r = [a_t^r,\ldots,a_{t+K-1}^r] \in \mathbb{R}^{K \times d_r}.
\label{eq:action-chunk}
\end{equation}
The VLM maps the visual language prefix to a sequence of $L_h$ hidden states:
\begin{equation}
    H_\phi = f_\phi(I_t^{1:V},\ell) \in \mathbb{R}^{L_h \times d_{\mathrm{vlm}}},
\label{eq:vlm-hidden}
\end{equation}
where $f_\phi$ denotes the VLM forward map, $L_h$ is the number of prefix tokens after visual and text tokenization, and $d_{\mathrm{vlm}}$ is the model dimension of the VLM and is 2560 in our setting.
A linear projection $\Pi$ maps $H_\phi$ to the DiT width.
The action expert is an 18-layer DiT with forward function $g_\theta$, width $1024$, $8$ attention heads, and head dimension $128$.
The current state and a noisy action chunk are projected with separate linear layers and concatenated as the DiT query sequence.
Our main configuration uses a separate DiT action expert with cross-attention to $\Pi(H_\phi)$, and can optionally interleave self-attention blocks in which the state and action tokens exchange information.
This differs from unified transformer VLA designs in which language, vision, state, and action tokens all share a single self-attention stack.

\paragraph{Embodiment agnostic batch format.}
State and action vectors are padded to $d_{\max}$, with an action valid mask $M^{\mathrm{act}}\in\{0,1\}^{K\times d_{\max}}$ over timestep and dimension.
$M^{\mathrm{act}}$ removes padded dimensions, padded frames, and annotation-only samples from the action losses.
Images are resized to a fixed resolution, and missing camera slots use dummy images with zero attention mask.
Dataset schemas specify state keys, action keys, camera mappings, action dimensions, gripper dimensions, and delta action masks, so single-arm, mobile manipulator, and dual arm embodiments share the same batch format.

\subsection{VLM Pretraining with FAST Tokens}\label{sec:vlm-pretrain}

Training a flow matching action head directly from a generic VLM is unstable because the VLM has never seen action semantics.
Without pretraining, the visual language prefix and the action chunk share no common representation.
We first tokenize continuous actions with FAST and train the VLM under next token supervision, so the prefix learns to predict action tokens before the DiT is attached.

In this stage, we do not instantiate the DiT.
The pipeline transforms continuous absolute action chunks with per-dimension dataset statistics, encodes them with the FAST tokenizer, and pads them to $d_{\max}$; padded action dimensions are never tokenized.
Depending on the schema, arm and gripper dimensions may use different statistics, such as mean--standard-deviation scaling for joints and $q_{01}/q_{99}$ alignment for compatible gripper endpoints.
States are normalized by the configured dataset statistics, clipped to $[-1,1]$, discretized into uniform bins, serialized as text, and prepended to the instruction.

Let $v_t$, $c_t$, $y_t$, and $z_{1:L_z}$ denote image tokens, state-conditioned instruction tokens, annotation tokens, and FAST action tokens, respectively, where $L_z$ is the FAST token sequence length.
The pretraining sequence is
\begin{equation}
    X_{\mathrm{pre}}
    =
    [v_t; c_t; y_t; z_{1:L_z}],
\label{eq:pretrain-sequence}
\end{equation}
where $c_t$ contains a discretized state string $b(q_t^r)$ (obtained by binning each robot state dimension into uniform bins and serializing the bin indices as text) prepended to the instruction.
The FAST objective is masked next token prediction:
\begin{equation}
    \mathcal{L}_{\mathrm{FAST}}
    =
    -\frac{1}{\sum_{i=1}^{L_z} m_i}
    \sum_{i=1}^{L_z}
    m_i \log p_\phi(z_i \mid v_t,c_t,y_t,z_{<i}).
\label{eq:fast-loss}
\end{equation}
Here $i$ indexes FAST token positions, $m_i$ masks padding tokens, and $p_\phi$ is the next token distribution induced by the VLM.
Samples that lack any action supervision (e.g., VQA or annotation only) skip the FAST block entirely instead of encoding an all-zero action.
If annotation targets are available, their token losses are added with weights $\lambda_j$, where $j$ indexes annotation targets:
\begin{equation}
    \mathcal{L}_{\mathrm{VLM}}
    =
    \mathcal{L}_{\mathrm{FAST}}
    +
    \sum_j \lambda_j \mathcal{L}^{(j)}_{\mathrm{CE}}.
\label{eq:vlm-loss}
\end{equation}

\subsection{Flow Matching Posttraining}\label{sec:flow-post}

Discrete FAST tokens align action semantics with the VLM prefix but lose the trajectory smoothness that laboratory manipulation requires.
The second stage therefore loads the VLM pretrained checkpoint, attaches the DiT action expert, and trains it with a flow matching objective~\cite{liu2022flow} that maps Gaussian noise to a clean action chunk through a deterministic vector field.

Given an active ground truth action chunk $A_t^r$, we pad it to $\widetilde{A}_t^r\in\mathbb{R}^{K\times d_{\max}}$ and sample Gaussian noise $\epsilon\sim\mathcal{N}(0,I)$ with the same shape.
For a flow time $\tau$, LabVLA forms
\begin{equation}
    \begin{aligned}
    X_\tau = \tau \widetilde{A}_t^r + (1-\tau)\epsilon,
    \qquad
    U_\tau = \widetilde{A}_t^r-\epsilon.
    \end{aligned}
\label{eq:flow-path}
\end{equation}
The implementation uses $\tau=0$ for noise and $\tau=1$ for clean actions, with $\tau=0.999\tilde{\tau}$ and $\tilde{\tau}\sim\mathrm{Beta}(1.0,1.5)$.
The DiT predicts
\begin{equation}
    V_\theta = g_\theta(X_\tau,\tau,q_t^r,\Pi(H_\phi)),
\label{eq:dit-velocity}
\end{equation}
where $\Pi$ is the VLM to DiT linear projection introduced after Eq.~(\ref{eq:vlm-hidden}).
The flow loss is the masked MSE
\begin{equation}
    \begin{aligned}
    S_M &= \sum_{k,d} M^{\mathrm{act}}_{k,d}, \\
    \mathcal{L}_{\mathrm{FM}}
    &=
    \begin{cases}
    S_M^{-1}\sum_{k,d} M^{\mathrm{act}}_{k,d}
    {(V_{\theta,k,d}-U_{\tau,k,d})}^{2}, & S_M>0,\\
    0, & S_M=0.
    \end{cases}
    \end{aligned}
\label{eq:flow-loss}
\end{equation}
Here $k$ and $d$ index action timesteps and dimensions.
Posttraining with knowledge insulation (KI, \Cref{sec:ki}) predicts absolute actions; task specific LabUtopia finetuning uses the same objective with delta action targets.
At sampling time the deterministic vector field reaches a usable trajectory in only $N{=}10$ Euler steps (Appendix~\ref{app:training-hyperparams}), well below the hundreds needed by diffusion policies and fast enough for closed loop laboratory control.

\subsection{Knowledge Insulation}\label{sec:ki}

Updating the VLM with the flow matching gradient drifts the linguistic and visual priors that aligned the prefix with laboratory instructions, degrading language following and visual grounding on rare instruments.
We therefore insulate the VLM from the flow loss while keeping the FAST and annotation token losses active, so the prefix can still learn from cross-entropy supervision without receiving velocity space gradients from the action expert.

Let $s_t=e_q(q_t^r)$ be the learned state token (with $e_q$ a linear embedding into VLM token space), and let $p_t$ denote the image and instruction prefix tokens, i.e.\ the $[v_t;c_t]$ block of Eq.~(\ref{eq:pretrain-sequence}) with the discretized state string replaced by $s_t$.
The VLM runs once over
\begin{equation}
    X_{\mathrm{KI}}
    =
    [s_t; p_t; y_t; z_{1:L_z}].
\label{eq:ki-sequence}
\end{equation}
Let $\operatorname{slice}_{p}$ select the hidden states at the prefix positions.
The DiT receives only the detached prefix slice,
\begin{equation}
    \begin{aligned}
    H_{\phi,p}^{\mathrm{KI}} &= \operatorname{slice}_{p}(f_\phi(X_{\mathrm{KI}})),\\
    \widetilde{H}_{\phi,p}^{\mathrm{KI}} &= \mathrm{sg}(H_{\phi,p}^{\mathrm{KI}}).
    \end{aligned}
\label{eq:detached-prefix}
\end{equation}
Here $\mathrm{sg}(\cdot)$ denotes stop-gradient.
Thus the KI velocity uses
\begin{equation}
    V_\theta^{\mathrm{KI}}
    =
    g_\theta(X_\tau,\tau,q_t^r,\Pi(\widetilde{H}_{\phi,p}^{\mathrm{KI}})).
\label{eq:ki-velocity}
\end{equation}
The flow loss updates the VLM to DiT projection and DiT, while token losses can still train the VLM\@.
The joint objective is
\begin{equation}
    \mathcal{L}_{\mathrm{KI}}
    =
    \alpha\mathcal{L}_{\mathrm{FM}}
    +
    \mathcal{L}_{\mathrm{FAST}}
    +
    \sum_j \lambda_j \mathcal{L}^{(j)}_{\mathrm{CE}},
    \qquad
    \alpha=10.
\label{eq:ki-loss}
\end{equation}
The FAST and annotation heads are training only and are removed at inference.
Knowledge insulation is a training time mechanism that blocks flow matching gradients from reaching the VLM prefix while FAST and annotation losses remain active.
In our setup, cotraining the VLM directly with the flow loss made the prefix representations less reliable for downstream attention.

\paragraph{Inference.}
At inference, LabVLA computes $H_\phi$ once, samples $X_0\sim\mathcal{N}(0,I)$, and integrates
\begin{equation}
    \begin{aligned}
    X_{\tau+\Delta\tau}
    =
    X_\tau
    +
    \Delta\tau\,g_\theta(X_\tau,\tau,q_t^r,\Pi(H_\phi)),
    \qquad
    \Delta\tau=1/N.
    \end{aligned}
\label{eq:euler-inference}
\end{equation}
Here $N$ is the number of Euler steps.
We output the first $K$ continuous actions, sliced to the action dimension of the active robot schema.
Hyperparameters for all three training phases (VLM pretraining, knowledge insulation posttraining, and LabUtopia finetuning) are listed in Appendix~\ref{app:training-hyperparams}.

%% file: section/experiment.tex
\section{Experiments}\label{sec:experiments}

\subsection{Experimental Setup}

\paragraph{Benchmark.}
We evaluate LabVLA on LabUtopia, which combines high fidelity simulation with procedural scene generation and a hierarchical task benchmark.
The setting matches our target because it requires labware recognition, contact rich manipulation, and instruction conditioned control.
The split covers six laboratory operations: picking up labware, pressing device buttons, opening doors, pouring liquids, heating beakers, and transporting beakers.
Each task is evaluated under an in-distribution (ID) setting that follows the training distribution of objects, layouts, and visual appearances, and an out-of-distribution (OOD) setting that perturbs object placement, appearance, or scene configuration to test generalization.
Every task is evaluated over 120 episodes per setting.

\paragraph{Baselines.}
We compare LabVLA against three families of recent VLA policies under the same LabUtopia protocol.
The sub 1B family includes SmolVLA~\cite{shukor2025smolvla} and X-VLA~\cite{zheng2025x}, which target affordable training.
The 3B family includes GR00T N1.5~\cite{bjorck2025gr00t}, $\pi_0$, $\pi_{0.5}$~\cite{intelligence2025pi_}, $\pi_0$-FAST, and InternVLA-A1~\cite{cai2026internvla}, spanning flow matching, FAST tokenized, and synthetic data pretrained policies.
The 4B family includes Wall-oss-flow~\cite{zhai2025igniting}.
All baselines are run from their public checkpoints under the LabUtopia evaluation harness, with action and state schemas adapted to the LabUtopia robot.

\subsection{Results}

\input{tables/main}

LabVLA achieves the highest average success rate among all evaluated baselines in both ID (71.1\%) and OOD (70.0\%), outperforming the next best policy $\pi_0$ by 7.8 and 6.8 percentage points respectively (\Cref{tab:level3_results}).
LabVLA leads on Pick Up (49.2\%/48.3\% ID/OOD) and Open Door (tied at 65.0\% ID, 65.8\% OOD), and scores 100\% on Press Button ID.
Press Button is near saturated for most baselines, so differences there are uninformative.
On Heat Beaker and Transport Beaker, GR00T N1.5 (99.2\%) and $\pi_{0.5}$ (90.0\% ID) lead respectively, while LabVLA stays competitive at 83.3\%--87.5\% and 85.8\%.
Pour Liquid remains the hardest category for all policies; no baseline exceeds 50\%, and liquid surface tracking remains unsolved.

LabVLA drops only 1.1\,pp from ID to OOD (71.1\%$\to$70.0\%), and its OOD average is the highest among all baselines.
The narrow gap suggests that domain randomization in LabEmbodied-Data builds visual and spatial invariances that transfer across scene perturbations.
Task difficulty tracks contact complexity rather than action horizon: Pour Liquid sits far below all others because tilting must be precise enough to avoid spilling and the policy receives no liquid level feedback, while multistep tasks like Heat Beaker and Transport Beaker are easier because their placement tolerances are generous.
Several baselines spike on individual tasks alongside near zero scores on others (e.g.\ SmolVLA 98.3\% Heat Beaker but 1.67\% Pour Liquid), whereas LabVLA is the most balanced, exceeding 48\% on every task except Pour Liquid (34.2\%).
For laboratory protocols that chain multiple operations in sequence, breadth across task families matters more than any single task peak.

\begin{figure*}[t]
    \centering
    \includegraphics[width=0.98\linewidth]{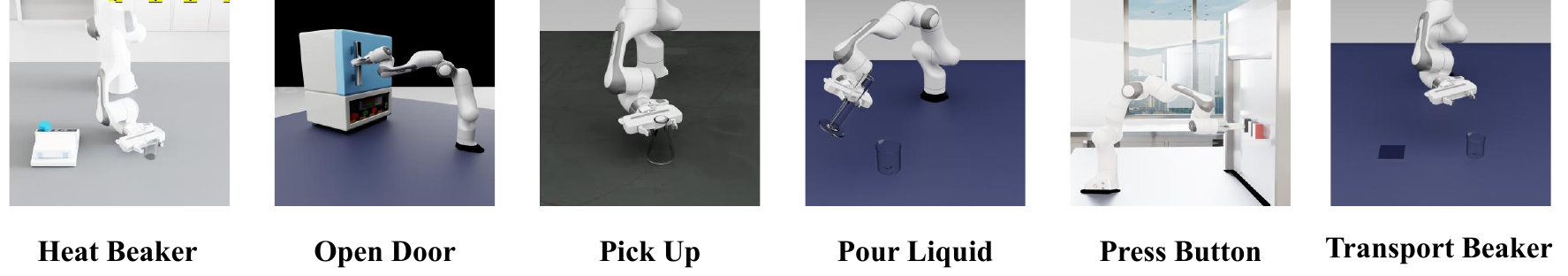}
    \caption{Representative rollout snapshots for the six LabUtopia evaluation tasks.
    Each column shows a third person view of the robot executing the task.
    \textbf{Heat Beaker}: the robot grasps a beaker and places it onto a heating plate.
    \textbf{Open Door}: the robot pulls open an equipment door by its handle.
    \textbf{Pick Up}: the robot picks up a target piece of labware from the bench.
    \textbf{Pour Liquid}: the robot tilts a source container to transfer liquid into a target vessel.
    \textbf{Press Button}: the robot locates and depresses a device button.
    \textbf{Transport Beaker}: the robot grasps a beaker, lifts it, and moves it to a designated location without dropping it.}
    \label{fig:task-vis}
\end{figure*}

\paragraph{Qualitative results.}
\Cref{fig:task-vis} shows representative rollout snapshots for the six LabUtopia evaluation tasks.
The tasks span a range of manipulation difficulty: Press Button requires only end effector positioning, while Pick Up and Open Door add grasp planning and articulated object interaction.
Heat Beaker and Transport Beaker are multistage tasks that require stable grasping, collision free transport, and precise placement onto a target surface.
Pour Liquid is the most demanding, requiring the policy to tilt a container at a controlled angle while maintaining a stable grip, consistent with its low success rates across all baselines in \Cref{tab:level3_results}.
LabVLA produces smooth, task appropriate trajectories across all six categories, though failure cases (not shown) typically involve premature grasp release during transport or imprecise tilt angles during pouring.

%% file: tables/main.tex
\begin{table}[t]
\centering
\caption{Success rates (\%) on LabUtopia tasks under in-distribution (ID) and out-of-distribution (OOD) settings.
\textbf{Bold} marks the column-best score.
The LabVLA size denotes the Qwen3-VL-4B-Instruct backbone together with the DiT action expert (Appendix~\ref{app:training-hyperparams}).}
\label{tab:level3_results}
\vspace{-2mm}
\renewcommand{\arraystretch}{1.18}
\resizebox{0.95\linewidth}{!}{%
\begin{tabular}{l c cccccc >{\centering\arraybackslash}p{3.2em}}
\toprule
\multirow{2}{*}{\textbf{Method}} & \multirow{2}{*}{\textbf{Size}} & \multicolumn{6}{c}{\textbf{LabUtopia Tasks}} & \multirow{2}{*}{\textbf{Avg.}} \\
\cmidrule(lr){3-8}
 & & Pick Up & Press Button & Open Door & Pour Liquid & Heat Beaker & Transport Beaker & \\
\midrule
\multicolumn{9}{l}{\textit{In-Distribution}} \\[2pt]
SmolVLA \cite{shukor2025smolvla}       & $<$1B & 15.8 & 97.5          & 16.7          & 0.8           & 96.7          & 85.8          & 52.2 \\
X-VLA \cite{zheng2025x}                & $<$1B & 27.5 & 98.3          & \textbf{65.0} & \textbf{45.0} & 25.8          & 83.3          & 57.5 \\
GR00T N1.5 \cite{bjorck2025gr00t}      & 3B    & 40.8 & 99.2          & 6.7           & 0             & \textbf{99.2} & 69.2          & 52.5 \\
$\pi_0$ \cite{black2024pi_0}           & 3B    & 21.7 & 92.5          & 51.6          & 37.5          & 90.0          & 86.7          & 63.3 \\
$\pi_{0.5}$ \cite{intelligence2025pi_} & 3B    & 38.3 & 60.0          & 55.8          & 29.2          & 40.8          & \textbf{90.0} & 52.4 \\
$\pi_0$-FAST \cite{pertsch2025fast}    & 3B    & 16.7 & 37.5          & 17.5          & 5.8           & 3.3             & 20.8          & 16.9 \\
InternVLA-A1 \cite{cai2026internvla}   & 3B    & 25.8 & 93.3          & 38.3          & 2.50          & 82.5          & 67.5          & 51.7 \\
Wall-oss-flow \cite{zhai2025igniting}  & 4B    & 11.7 & 54.2          & 0.83          & 0             & 0             & 29.2          & 16.0 \\

\rowcolor{blue!6}
\textbf{LabVLA}                    & 5B    & \textbf{49.2}   & \textbf{100}  & \textbf{65.0}   & 43.3            & 83.3            & 85.8          & \textbf{71.1} \\
\midrule
\multicolumn{9}{l}{\textit{Out-of-Distribution}} \\[2pt]
SmolVLA \cite{shukor2025smolvla}       & $<$1B & 11.7 & \textbf{99.2} & 18.3          & 1.67          & 98.3          & \textbf{89.2} & 53.1 \\
X-VLA \cite{zheng2025x}                & $<$1B & 27.5 & \textbf{99.2} & 59.2          & 25.0          & 39.2          & 67.5          & 52.9 \\
GR00T N1.5 \cite{bjorck2025gr00t}      & 3B    & 33.3 & 92.5          & 8.3           & 0             & \textbf{99.2} & 66.7          & 50.0 \\
$\pi_0$ \cite{black2024pi_0}           & 3B    & 19.2 & 89.1          & 53.3          & \textbf{38.3} & 90.8          & 88.3          & 63.2 \\
$\pi_{0.5}$ \cite{intelligence2025pi_} & 3B    & 30.0 & 68.3          & 59.2          & 29.2          & 40.0          & 85.8          & 52.1 \\
$\pi_0$-FAST \cite{pertsch2025fast}    & 3B    & 14.2 & 45.0          & 15.8          & 7.5           & 11.7          & 24.2          & 19.7 \\
InternVLA-A1 \cite{cai2026internvla}   & 3B    & 19.2 & 95.8          & 63.3          & 0.83          & 84.2          & 57.5          & 53.5 \\
Wall-oss-flow \cite{zhai2025igniting}  & 4B    & 7.50 & 61.7          & 0             & 0             & 0             & 26.7          & 16.0 \\

\rowcolor{blue!6}
\textbf{LabVLA}                    & 5B    & \textbf{48.3}   & 98.3          & \textbf{65.8}   & 34.2            & 87.5            & 85.8          & \textbf{70.0} \\
\bottomrule
\end{tabular}%
}
\end{table}

%% file: section/analysis.tex
\section{Analysis}\label{sec:analysis}

We present two followup analyses beyond the main LabUtopia comparison: one testing whether LabEmbodied-Data benefits an external policy, and one testing simulation to real transfer on physical hardware.

\input{tables/analysis2}

\subsection{LabEmbodied-Data Transferability}\label{sec:analysis_xvla_transfer}

To test whether LabEmbodied-Data is useful beyond the LabVLA architecture, we finetune X-VLA~\cite{zheng2025x}, a sub 1B baseline already included in the LabUtopia comparison, on LabEmbodied-Data and evaluate it on the same five nonsaturated LabUtopia tasks (Press Button is excluded because it is near saturated for all baselines).
\Cref{tab:analysis_xvla_transfer} shows that adding LabEmbodied-Data improves X-VLA on four of five ID tasks and all five OOD tasks.
The five task average rises from 49.3\% to 64.3\% ID (+15.0\,pp) and from 43.7\% to 63.0\% OOD (+19.3\,pp).
The largest gains appear on Heat Beaker (ID: 25.8\%$\to$68.3\%) and Pour Liquid (OOD: 25.0\%$\to$65.0\%), both involving instrument specific contact patterns absent from X-VLA's original training data.
Pick Up is the only task where the augmented model does not improve in-distribution (27.5\%$\to$26.7\%), though it does improve OOD (27.5\%$\to$31.7\%).
LabEmbodied-Data therefore provides transferable laboratory supervision not specific to the LabVLA architecture.

\subsection{Real world Experiments}\label{sec:analysis_franka}

\begin{wrapfigure}{r}{0.42\linewidth}
    \vspace{-3mm}
    \centering
    \includegraphics[width=\linewidth]{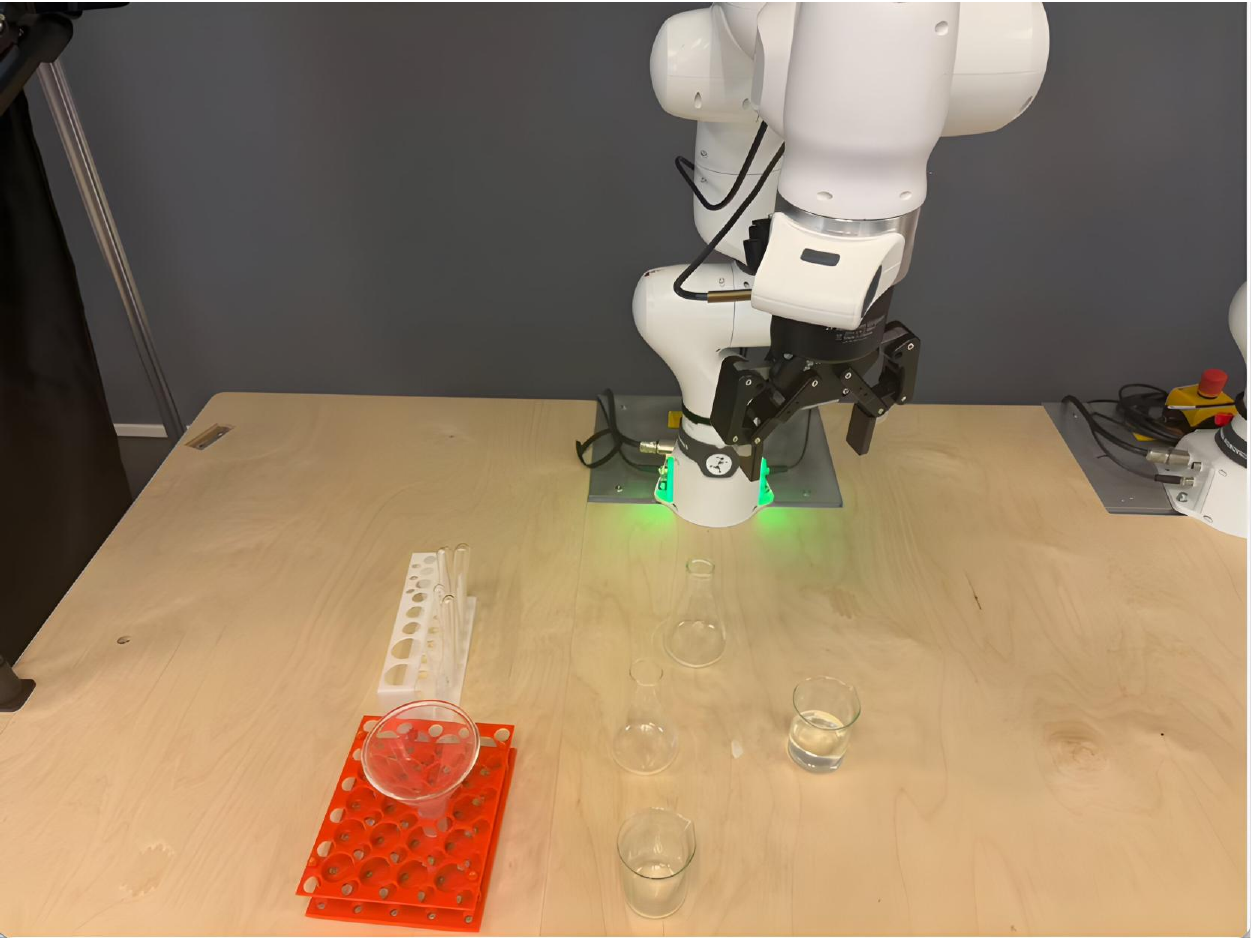}
    \caption{The Franka platform and laboratory workspace used for real robot experiments, with representative labware arranged on the bench.}
    \label{fig:real_world}
    \vspace{-4mm}
\end{wrapfigure}

We deploy LabVLA on a physical Franka platform alongside DreamZero~\cite{ye2026world} and $\pi_{0.5}$~\cite{intelligence2025pi_} to test simulation to real transfer.
We design four tasks, each composing 2--4 atomic laboratory skills:
\textbf{Shake Liquid} (pick$\to$shake$\to$place),
\textbf{Pour Liquid} (pick$\to$pour$\to$place),
\textbf{Magnetic Stir} (pick$\to$place$\to$press, operating a magnetic stirrer), and
\textbf{Funnel Plug/Unplug} (pick$\to$place$\to$pick$\to$place).
For each task we collect 50 demonstrations in which the target object and its final placement are randomized within a $5\times5$\,cm region.

Each policy is evaluated under four conditions crossing target position (in domain vs.\ out of domain) and workspace clutter (clean vs.\ cluttered).
\Cref{tab:real_robot} reports success rates.
All three policies exceed 70\% in most conditions; simulation pretraining transfers to physical hardware.
On the four task average, DreamZero slightly outperforms LabVLA in cluttered settings (81.0\% vs.\ 80.0\% in domain; 75.5\% vs.\ 74.0\% out of domain), while the two are within 0.5\,pp in clean in domain (87.0\% vs.\ 86.5\%) and LabVLA leads by 2.0\,pp in clean out of domain (80.0\% vs.\ 78.0\%).
The gap between the two is within run to run variance for most individual tasks.
Both consistently outperform $\pi_{0.5}$, whose average drops to 71.5\% under cluttered out of domain conditions.
Pour Liquid degrades the most under clutter and position shift for all policies (86\%$\to$72\% for LabVLA, 88\%$\to$70\% for DreamZero); precise liquid transfer is sensitive to both factors.
Funnel Plug/Unplug, the longest horizon task at four skill steps, is the hardest overall; LabVLA reaches 80\% in clean out of domain, ahead of both baselines in that condition.

\input{tables/real_robot}

%% file: tables/analysis2.tex
\begin{table}[h]
\centering
\caption{Transferability of LabEmbodied-Data to an external policy.
We fine-tune X-VLA~\cite{zheng2025x} on LabEmbodied-Data and evaluate on five LabUtopia tasks under in-distribution (ID) and out-of-distribution (OOD) settings.
Adding LabEmbodied-Data lifts the five-task average by +15.0\,pp (ID) and +19.3\,pp (OOD), with the largest gains on tasks requiring instrument specific contact patterns.}
\label{tab:analysis_xvla_transfer}
\vspace{-2mm}
\renewcommand{\arraystretch}{1.18}
\resizebox{0.95\linewidth}{!}{%
\begin{tabular}{l c ccccc >{\centering\arraybackslash}p{3.2em} >{\centering\arraybackslash}p{3.2em}}
\toprule
\multirow{2}{*}{\textbf{Method}} & \multirow{2}{*}{\textbf{Size}} & \multicolumn{5}{c}{\textbf{LabUtopia Tasks}} & \multirow{2}{*}{\textbf{Avg.}} & \multirow{2}{*}{$\boldsymbol{\Delta}$} \\
\cmidrule(lr){3-7}
 & & Pick Up & Open Door & Pour Liquid & Heat Beaker & Transport Beaker & & \\
\midrule
\multicolumn{9}{l}{\textit{In-Distribution}} \\[2pt]
X-VLA \cite{zheng2025x} & $<$1B & 27.5 & 65.0 & 45.0 & 25.8 & 83.3 & 49.3 & --- \\
X-VLA + LabEmbodied & $<$1B & 26.7 & 69.2 & 59.2 & 68.3 & 98.3 & 64.3 & \textbf{+15.0} \\
\midrule
\multicolumn{9}{l}{\textit{Out-of-Distribution}} \\[2pt]
X-VLA \cite{zheng2025x} & $<$1B & 27.5 & 59.2 & 25.0 & 39.2 & 67.5 & 43.7 & --- \\
X-VLA + LabEmbodied & $<$1B & 31.7 & 63.3 & 65.0 & 65.0 & 90.0 & 63.0 & \textbf{+19.3} \\
\bottomrule
\end{tabular}%
}
\end{table}

%% file: tables/real_robot.tex
\begin{table*}[h]
\centering
\small
\caption{Real robot evaluation on a Franka platform.
Each task composes 2--4 atomic skills; demonstrations are collected with target objects randomly placed within a $5\times5$\,cm region.
\textbf{In-domain}: target positions drawn from the training distribution.
\textbf{Out-of-domain}: target positions outside the training region.
\textbf{Clutter}: additional distractor objects placed in the workspace.
Success rate (\%) is reported over 50 rollouts per setting.
\textbf{Bold} marks the row best.}\label{tab:real_robot}
\resizebox{\textwidth}{!}{%
\begin{tabular}{l l c c c c}
\toprule
Task & Location & Clutter & LabVLA (Ours) & DreamZero~\cite{ye2026world} & $\pi_{0.5}$~\cite{intelligence2025pi_} \\
\midrule
\multirow{4}{*}{\parbox{4.0cm}{Shake Liquid\\[-1pt]{\scriptsize pick\,$\to$\,shake\,$\to$\,place}}}
 & \multirow{2}{*}{In-domain}     & \ding{55} & \textbf{92} & 90 & \textbf{92} \\
 &                                 & \ding{51} & \textbf{86} & 84 & 80 \\
 & \multirow{2}{*}{Out-of-domain} & \ding{55} & \textbf{84} & \textbf{84} & 82 \\
 &                                 & \ding{51} & \textbf{80} & \textbf{80} & 78 \\
\midrule
\multirow{4}{*}{\parbox{4.0cm}{Pour Liquid\\[-1pt]{\scriptsize pick\,$\to$\,pour\,$\to$\,place}}}
 & \multirow{2}{*}{In-domain}     & \ding{55} & 86 & \textbf{88} & 82 \\
 &                                 & \ding{51} & 78 & \textbf{80} & 74 \\
 & \multirow{2}{*}{Out-of-domain} & \ding{55} & \textbf{76} & 72 & 74 \\
 &                                 & \ding{51} & \textbf{72} & 70 & 68 \\
\midrule
\multirow{4}{*}{\parbox{4.0cm}{Magnetic Stir\\[-1pt]{\scriptsize pick\,$\to$\,place\,$\to$\,press}}}
 & \multirow{2}{*}{In-domain}     & \ding{55} & \textbf{88} & 86 & \textbf{88} \\
 &                                 & \ding{51} & 80 & \textbf{84} & 80 \\
 & \multirow{2}{*}{Out-of-domain} & \ding{55} & 80 & 78 & \textbf{82} \\
 &                                 & \ding{51} & 74 & \textbf{80} & 76 \\
\midrule
\multirow{4}{*}{\parbox{4.0cm}{Funnel Plug/Unplug\\[-1pt]{\scriptsize pick\,$\to$\,place\,$\to$\,pick\,$\to$\,place}}}
 & \multirow{2}{*}{In-domain}     & \ding{55} & 80 & \textbf{84} & 78 \\
 &                                 & \ding{51} & \textbf{76} & \textbf{76} & 72 \\
 & \multirow{2}{*}{Out-of-domain} & \ding{55} & \textbf{80} & 78 & 70 \\
 &                                 & \ding{51} & 70 & \textbf{72} & 64 \\
\midrule
\multirow{4}{*}{Average}
 & \multirow{2}{*}{In-domain}     & \ding{55} & 86.5 & \textbf{87.0} & 85.0 \\
 &                                 & \ding{51} & 80.0 & \textbf{81.0} & 76.5 \\
 & \multirow{2}{*}{Out-of-domain} & \ding{55} & \textbf{80.0} & 78.0 & 77.0 \\
 &                                 & \ding{51} & 74.0 & \textbf{75.5} & 71.5 \\
\bottomrule
\end{tabular}}
\end{table*}

%% file: section/related_work.tex
\section{Related Work}\label{sec:related-work}

\subsection{Vision-Language-Action Models}

\paragraph{Mainstream VLA training and grounding.}
Mainstream VLA research learns robot policies that generate actions from visual observations and language instructions.
CLIPort~\cite{shridhar2022cliport} grounded pick-and-place with visual language representations, BC-Z~\cite{jang2022bc} studied zero-shot task generalization through language conditioned imitation learning, and RT-1~\cite{brohan2022rt} and RT-2~\cite{zitkovich2023rt} scaled Transformer policies with large robot datasets and web-scale vision language pre-training.
OpenVLA~\cite{kim2024openvla} and CogACT~\cite{li2024cogact} developed open VLA training and a cognition-action separation, while $\pi_0$, $\pi_{0.5}$, OpenVLA-OFT~\cite{kim2025fine}, and $\pi_0$-FAST improved flow based action generation, FAST tokenization, and faster decoding.
Recent baselines targeted by our experiments extend this line in several directions: X-VLA with cross embodiment soft prompts, GR00T N1.5 with generalist humanoid pretraining, InternVLA-A1 with unified understanding, generation, and action modeling, Wall-oss-flow with embodied space VLM ignition, and UniVLA~\cite{bu2025univla} with large scale latent action learning.
These methods establish VLA training on household and tabletop demonstrations; LabVLA carries the same training principles into protocol conditioned laboratory data, where the supervision distribution rather than the policy class is the dominant variable.

\paragraph{Efficiency oriented VLA.}
Efficiency-oriented models reduce model or inference cost through compact architectures, dynamic inference, state-space models, or diffusion policies.
SmolVLA and TinyVLA~\cite{wen2025tinyvla} target affordable or data efficient VLA training, while DeeR-VLA~\cite{yue2024deer}, RoboMamba~\cite{liu2024robomamba}, and RDT-1B~\cite{liu2025rdt} explore dynamic inference, efficient sequence modeling, and diffusion based action modeling.
The efficiency mechanisms in this line are orthogonal to LabVLA; we focus on data distribution and training recipe rather than parameter or latency reduction.

\paragraph{Reasoning, memory, and spatial VLA.}
A growing body of work improves generalization by augmenting VLA models with reasoning traces, memory, or explicit spatial representations.
Robotic chain-of-thought~\cite{zawalski2024robotic}, CoT-VLA~\cite{zhao2025cot}, CoA-VLA~\cite{li2025coa}, and ThinkAct~\cite{huang2026thinkact} use intermediate reasoning or latent planning to guide robot actions, while FlowVLA~\cite{zhong2025flowvla} and InstructVLA~\cite{yang2025instructvla} study motion reasoning and instruction tuning, and MemoryVLA~\cite{shi2025memoryvla}, MAP-VLA~\cite{li2025map}, and TraceVLA~\cite{zheng2025tracevla} address temporal context and demonstration retrieval.
Spatial models use voxel, view transformer, feature field, pixel level, or 3D aware representations: Perceiver-Actor~\cite{shridhar2023perceiver}, RVT~\cite{goyal2023rvt}, RVT-2~\cite{goyal2024rvt}, GNFactor~\cite{ze2023gnfactor}, Act3D~\cite{gervet2023act3d}, SpatialVLA~\cite{qu2025spatialvla}, and PixelVLA~\cite{liang2025pixelvla} all inject spatial or pixel level information directly into VLA models.
These augmented representations target broader generalization on tabletop data; laboratory protocols introduce a complementary generalization axis, namely instrument and physical state diversity, that these methods do not directly address.

\paragraph{Cross-embodiment datasets.}
Large scale datasets such as Open X-Embodiment, DROID, and BridgeData V2 aggregate real world cross embodiment demonstrations to support general VLA training.
These corpora are essential supervision for everyday manipulation, but they do not include laboratory instruments or protocol conditioned trajectories; LabVLA complements them with success filtered protocol conditioned simulation rather than competing with them on coverage of household and tabletop data.

\subsection{Scenarios and Benchmarks}

Robot learning benchmarks are defined by their scene distribution as much as by their policy class.
Tabletop and simulated manipulation suites~\cite{yu2020meta,zhu2020robosuite,james2020rlbench,gu2023maniskill2,jiang2023vima,mees2022calvin,liu2023libero} stress multitask control, language following, long horizon execution, and knowledge transfer; real robot datasets~\cite{dasari2019robonet,bu2025agibot} extend these settings through broader embodiments and collection sites.
Household and room scale benchmarks~\cite{li2021igibson,shridhar2020alfred,szot2021habitat,li2023behavior,yenamandra2023homerobot,nasiriany2024robocasa} make scene context central through object distributions, furniture constraints, task horizons, and evaluation protocols, while scene specific suites such as SoftGym~\cite{lin2021softgym}, FurnitureBench~\cite{heo2025furniturebench}, and FMB~\cite{luo2025fmb} isolate nontabletop skills like deformable object manipulation, furniture assembly, and functional manipulation.

Scientific laboratories form a distinct scene family because the robot must interact with instruments, containers, materials, and protocol state.
Earlier self-driving laboratory work built specialized hardware platforms for closed loop experiment planning and chemical discovery, including organic synthesis~\cite{granda2018controlling}, literature to execution translation~\cite{mehr2020universal}, mobile robotic chemists~\cite{burger2020mobile}, and autonomous materials synthesis platforms~\cite{szymanski2023autonomous}.
Simulation based laboratory benchmarks~\cite{li2024chemistry3d} and chemistry oriented planning systems~\cite{yoshikawa2022chemistry,huang2025tarmac} further show that laboratory protocols require constrained planning over instruments, materials, and skills.
LabVLA differs from both lines: unlike prior VLAs trained on household or tabletop data, and unlike prior laboratory simulators such as LabUtopia and Chemistry3D that ship a benchmark without a paired policy, it couples a protocol conditioned data engine (RoboGenesis) with a Qwen3-VL based policy trained via FAST pretraining and flow matching posttraining under a single cross embodiment action observation schema.
The contribution is the connection between executable laboratory workflows, success filtered demonstrations, and continuous action VLA learning, not any single component in isolation.

%% file: section/conclusion.tex
\section{Conclusion}\label{sec:conclusion}

This paper studies VLA learning for scientific laboratory automation.
Our main claim is that data and embodiment are the primary bottlenecks alongside model design: a written scientific protocol becomes executable only when scenes, instruments, physical state, and robot morphologies share a usable supervision schema.
LabVLA operationalizes this claim by pairing RoboGenesis synthesized LabEmbodied-Data with a Qwen3-VL policy trained through FAST action token pretraining and flow matching posttraining under a knowledge insulation design.
Under the LabUtopia harness, LabVLA achieves the highest average success rate among all evaluated baselines in both ID and OOD settings, with Pour Liquid remaining the main open category.
A four task study on a physical Franka platform further shows that simulation pretraining transfers to real benchtop manipulation.

We provide RoboGenesis (a programmable workflow and data engine), LabEmbodied-Data (an annotated laboratory corpus), and the LabVLA training recipe as reusable artifacts.
Other groups can extend them with new instruments, protocols, and robots without redoing the underlying infrastructure, which lowers the entry cost for laboratory VLA research.
Next steps are to move beyond benchtop evaluation toward deployment in working laboratories with real reagents and instruments under explicit safety constraints, and to scale RoboGenesis to broader wet chemistry and biology workflows.

%% file: section/discussion.tex
\section{Discussion}\label{sec:discussion}

\paragraph{What the current results support.}
The LabUtopia results in \Cref{tab:level3_results} support a stratified rather than averaged view of laboratory manipulation.
Press Button is near saturated (most baselines $\geq$92\%, LabVLA 100\%/98.3\% ID/OOD), and Heat Beaker is solved by the strongest baselines (GR00T N1.5 99.2\%, SmolVLA 96.7\%/98.3\%) though not uniformly across all methods.
Operations that add contact stability, container geometry, and longer action horizons, namely picking, opening, pouring, and transport, spread the methods further apart, with Pour Liquid (LabVLA 43.3\%/34.2\%) sitting far below the device interaction tasks.
This separation exposes which parts of simulated laboratory protocol execution are already approachable and which parts still need better data, control, or recovery.

\begin{wrapfigure}{r}{0.42\linewidth}
    \vspace{-3mm}
    \centering
    \includegraphics[width=\linewidth]{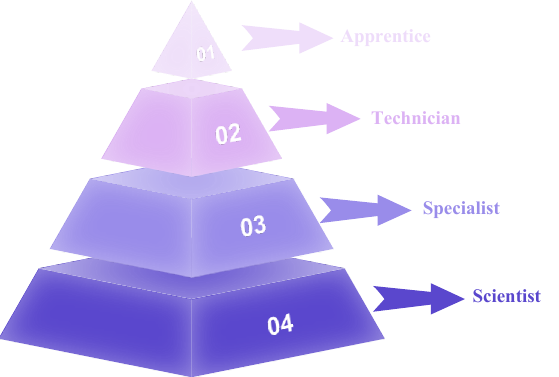}
    \caption{Four-tier capability pyramid for laboratory VLA: from \hlalt{Apprentice} to \hlalt{Scientist}.}
    \label{fig:tiers}
    \vspace{-4mm}
\end{wrapfigure}

\paragraph{Levels of embodied laboratory competence.}
Rather than a single aggregate score, laboratory manipulation is better viewed through four levels of competence modeled on real laboratory roles.
Level~1 (\emph{Apprentice}) covers single step interactions with laboratory objects: grasping labware, pressing a button, opening a door, or placing a container.
Level~2 (\emph{Technician}) requires following a written multistep protocol through physical state changes such as pouring, heating, stirring, shaking, or transporting a vessel, where a failed earlier step cascades through the rest of the procedure.
Level~3 (\emph{Specialist}) adds operation of precision instruments (pipettes, centrifuges, thermal cyclers, microscopes) in longer workflows with measurement logging and safety constraints.
Level~4 (\emph{Scientist}) modifies the procedure in response to observations or measurements: adjusting concentrations, branching to alternative protocols, or deciding when an experimental objective has been met.
We position LabVLA at Level~2 (Technician).
LabVLA can execute several fixed simulated protocols, and RoboGenesis can express multistep laboratory workflows with step level annotations.
However, the policy does not yet demonstrate the instrument competence, measurement awareness, or scientific judgment that Level~3 and Level~4 require.

\paragraph{Limitations.}
LabVLA is an early study of VLA learning for scientific laboratory settings, and several gaps remain between what we demonstrate and what real laboratory deployment would require.
First, most of our validation happens inside a simulated bench environment; the real robot study covers four benchtop tasks on a single Franka platform.
Real scientific laboratories add hardware drift, reagent variability, safety constraints, contamination risks, and unanticipated failure modes that no sanitized benchmark captures, so the gap between these numbers and a system one would trust on an actual wet lab experiment is substantial.
Second, the autonomy we demonstrate is \emph{protocol following} only: given a fixed procedure, LabVLA can attempt to execute it.
It does not yet choose experimental conditions, revise a protocol from measurements, substitute reagents, or decide when a scientific objective has been satisfied.
Third, even within today's scope, we have not yet shown that a protocol trained policy can collaborate naturally with human scientists, communicate intermediate observations, or recognize and explain its own failure modes.
An embodied scientific assistant will need these properties before it can be trusted inside a real laboratory loop.
Closing the distance between simulated protocol execution and real embodied laboratory practice will take considerably more work than one paper can cover.

\paragraph{From AI for research to embodied AI for science.}
We view this work as one step toward giving AI for research a grounded interface to laboratory work.
Today's AI for research already reads literature, writes code, designs hypotheses, and plans experiments, but execution still falls to a human operator who picks up the pipette, sets the heater, and watches for the color change.
RoboGenesis and LabVLA outline one practical path: a simulation based workflow and data engine that captures structured laboratory procedures, paired with a VLA training recipe for fixed protocol execution.
As such engines mature, embodied AI may assist with routine parts of protocol execution under human supervision, from reagent preparation to simple instrument operation and observation logging.
The longer term goal is not to replace scientific judgment, but to let scientific AI systems contribute to parts of experimental practice while keeping human oversight over hypotheses, safety, and interpretation.

%% file: section/appendix.tex
\newpage

\section{Training Hyperparameters}\label{app:training-hyperparams}

\begin{table}[h]
\centering
\small
\caption{Implementation and training hyperparameters used for LabVLA\@.
VLM pretraining and KI posttraining use absolute action targets; finetuning uses delta action targets.
``KI posttraining'' denotes the flow matching stage with knowledge insulation enabled.}\label{tab:training-hyperparams}
\begin{tabular}{>{\raggedright\arraybackslash}p{0.25\linewidth}>{\raggedright\arraybackslash}p{0.20\linewidth}>{\raggedright\arraybackslash}p{0.20\linewidth}>{\raggedright\arraybackslash}p{0.20\linewidth}}
\toprule
Hyperparameter & VLM pretraining & KI posttraining & Finetuning \\
\midrule
Image resolution & $224\times224$ & $224\times224$ & $224\times224$ \\
Action horizon $K$ & 50 & 50 & 50 \\
Maximum state/action dimension $d_{\max}$ & 32 / 32 & 32 / 32 & 32 / 32 \\
State discretization bins & 256 & N/A & N/A \\
Inference Euler steps $N$ & N/A & 10 & 10 \\
Optimizer & AdamW & AdamW & AdamW \\
Batch size per GPU & 64 & 64 & 48 \\
Global batch size & 1536 & 1024 & 192 \\
Gradient accumulation & 1 & 1 & 1 \\
Training GPUs & 24$\times$A100 & 16$\times$A100 & 4$\times$A100 \\
Training steps & 100k & 80k & 80k \\
VLM learning rate & $5\times10^{-5}$ & $1\times10^{-5}$ & $5\times10^{-5}$ \\
DiT/action learning rate & N/A & $10^{-4}$ & $5\times10^{-5}$ \\
Weight decay & 0.01 & 0.01 & 0.01 \\
Gradient clipping & 1.0 & 1.0 & 1.0 \\
Warmup steps & 1k & 8k & 2k \\
Final decay learning rate & $2.5\times10^{-6}$ & $5\times10^{-7}$ & $5\times10^{-5}$ \\
Attention backend & FlashAttention-2 & FlashAttention-2 & FlashAttention-2 \\
Distributed training & DeepSpeed ZeRO-2 & DeepSpeed ZeRO-2 & DeepSpeed ZeRO-2 \\
\midrule
\multicolumn{4}{l}{\textit{Model size.}\quad VLM backbone: Qwen3-VL-4B-Instruct.\quad DiT: 18 layers, width 1024, 8 heads, head dim 128.} \\
\bottomrule
\end{tabular}
\end{table}

\section{Training History}\label{app:training-lessons}

We document several failure modes encountered during LabVLA development that are specific to multidataset VLA training and not easily diagnosed from training loss alone.
These observations come from our training logs and closed loop evaluation records.

\paragraph{Action dimension padding implicitly rescales gradients.}
LabVLA pads all action vectors to a fixed 32 dimensional tensor, of which typically only 8 dimensions are active for a single arm robot.
Early runs averaged the flow matching loss over all 32 dimensions; padded dimensions contribute zero error but still count in the denominator, which silently scales the loss, and hence the action expert gradient, down by $4\times$.
The final recipe instead slices the per element loss to each robot's active action dimensions and averages only over those dimensions and non padded frames, so embodiments with different degrees of freedom receive the same per element gradient scale.
Making this switch is mathematically equivalent to raising the action expert learning rate by $4\times$, and finetuning performance degraded until the learning rate was retuned to compensate.
We therefore treat action dimensionality and mask normalization changes as optimizer changes, not bookkeeping fixes, and revalidate any such modification by closed loop evaluation.

\paragraph{Warmstart base quality dominated architecture choices.}
The largest single factor in our TransportBeaker finetuning was the quality and data composition of the warmstart checkpoint, not the DiT architecture or the learning rate schedule.
Under the same finetuning recipe, increasing the diversity and volume of the posttraining data improved TransportBeaker success from 60\% to 86\% on a 120-episode evaluation.
We attribute this to the action prior learned during posttraining: the larger mixture included high fidelity real robot demonstrations covering grasping distributions closer to the target task, whereas the smaller base had broader scene diversity but weaker contact level action coverage.

\section{Memory and Compute Optimizations}\label{app:training-efficiency}

We describe the engineering measures that made large batch training of the Qwen3-VL-4B backbone plus DiT stack ($\approx$5B parameters in total) practical on 80\,GB GPUs.
These are implementation choices for our specific configuration and should not be interpreted as hardware independent benchmarks.

\paragraph{Selective gradient checkpointing across submodules.}
LabVLA exposes gradient checkpointing as three independent flags covering the vision encoder, the language model, and the DiT action head.
To balance memory against speed, training enables only the language model flag: the LM processes the full merged multimodal sequence through the deepest and widest layers and dominates activation memory, so checkpointing it alone frees most of the memory that full checkpointing would while recomputing only one submodule, and this is what lets the production batch size fit on 80\,GB GPUs.
Combined with the fused kernels, background batch prefetching, and host memory management described below, these measures successfully reduced per step training time.

\paragraph{Fused GPU kernels for the VLM backbone.}
LabVLA applies Liger-Kernel~\cite{hsu2024liger} fused operators to the Qwen3-VL backbone at four sites: RMSNorm, RoPE, SwiGLU, and the annotation cross-entropy loss.
Fused RMSNorm eliminates the fp32 upcasting overhead of the standard implementation; fused RoPE reduces intermediate sin/cos tensor allocation.
For SwiGLU, the upstream Liger integration call is a known no-op on versions $\leq$0.7.x, so we version gate a manual per layer MLP rebind that applies the fused kernel directly; versions $\geq$0.8 are assumed to support native patching.
The most critical optimization is fused linear cross-entropy (FLCE) for annotation supervision.
The standard path materializes a dense logits tensor of shape $(B, L, V)$ where $V \approx 152\text{K}$ is the vocabulary size; at batch size 64 and annotation length 256, this tensor alone consumes roughly 10\,GB of fp32 memory and does not fit alongside the 4B VLM backbone on an 80\,GB GPU.
FLCE chunk fuses the weight projection, softmax, and cross-entropy computation so the full logits tensor is never allocated.
Because a missing Liger installation would otherwise crash only upon the first annotation batch (potentially tens of minutes into a distributed job), we validate the FLCE import at training startup before any GPU memory is allocated.

\paragraph{Attention mask design.}
The training sequence concatenates image, instruction, annotation, and FAST action tokens.
The production configuration applies standard decoder-only causal attention over this sequence with a one dimensional padding mask, the layout FlashAttention-2 consumes directly: every token attends to all earlier non padded positions.
We also implemented the blockwise mask prescribed by $\pi_{0.5}$, in which the image and instruction prefix forms one fully bidirectional block, annotation tokens attend to the full prefix, FAST tokens attend to the prefix and annotations, each block remains causal within itself, and the prefix never attends forward.
That pattern requires an arbitrary 4D additive mask, which FlashAttention-2 does not support; routing it through SDPA instead raised step time by roughly 30\%.

\paragraph{Background batch prefetch with a dedicated transfer stream.}
Even with fast kernels, a training step stalls whenever the GPU waits for the DataLoader or for the host to GPU copy of the next batch.
The training loop therefore runs a producer thread that pulls batches from the DataLoader, moves each one to the GPU on a dedicated CUDA transfer stream, and parks the result in a small bounded queue that is filled to 80\% before the first step.
Each queued batch carries a CUDA event recorded on the transfer stream, and the compute stream waits only on that event at consumption time, so the copy of the next batch overlaps with the compute of the current one.
With the buffer warm, data loading and transfer leave the critical path and the step time approaches pure compute time.

\paragraph{Host memory management for multiadapter video caching.}
Streaming training decodes frames on demand from compressed MP4 shards through PyAV, and reopening a container for every read is prohibitively slow, so each dataset adapter keeps an LRU cache of open containers bounded at 64.
The 4 dataset mixture instantiates over 60 adapter shards per DataLoader worker, and persistent workers (required for homogeneous mixture batching) let these individually bounded caches compose to $60 \times 64 = 3{,}840$ potentially open containers per process, each holding codec state and buffered packets.
The deeper problem is allocator behavior rather than the cache bound itself: decoder allocations of mixed sizes interleave across glibc's per thread malloc arenas, so even after eviction closes a container, the freed chunks fragment the arenas and \texttt{free()} rarely returns pages to the operating system.
Worker resident set size therefore climbed toward 42\,GB while live allocations stayed bounded.
We made the bound global rather than per adapter by replacing the independent caches with one process wide shared LRU (capacity 256), capped the arena count with \texttt{MALLOC\_ARENA\_MAX=2} so freed decoder buffers coalesce, and added periodic \texttt{malloc\_trim(0)} calls in both the main process and the workers to return free heap pages to the operating system.
Together these reduced mean worker RSS by 55\% at a 3--8\% step time cost from the extra container reopens.

\section{Case Study}\label{app:case-study}

\begin{figure*}[h]
    \centering
    \includegraphics[width=0.98\linewidth]{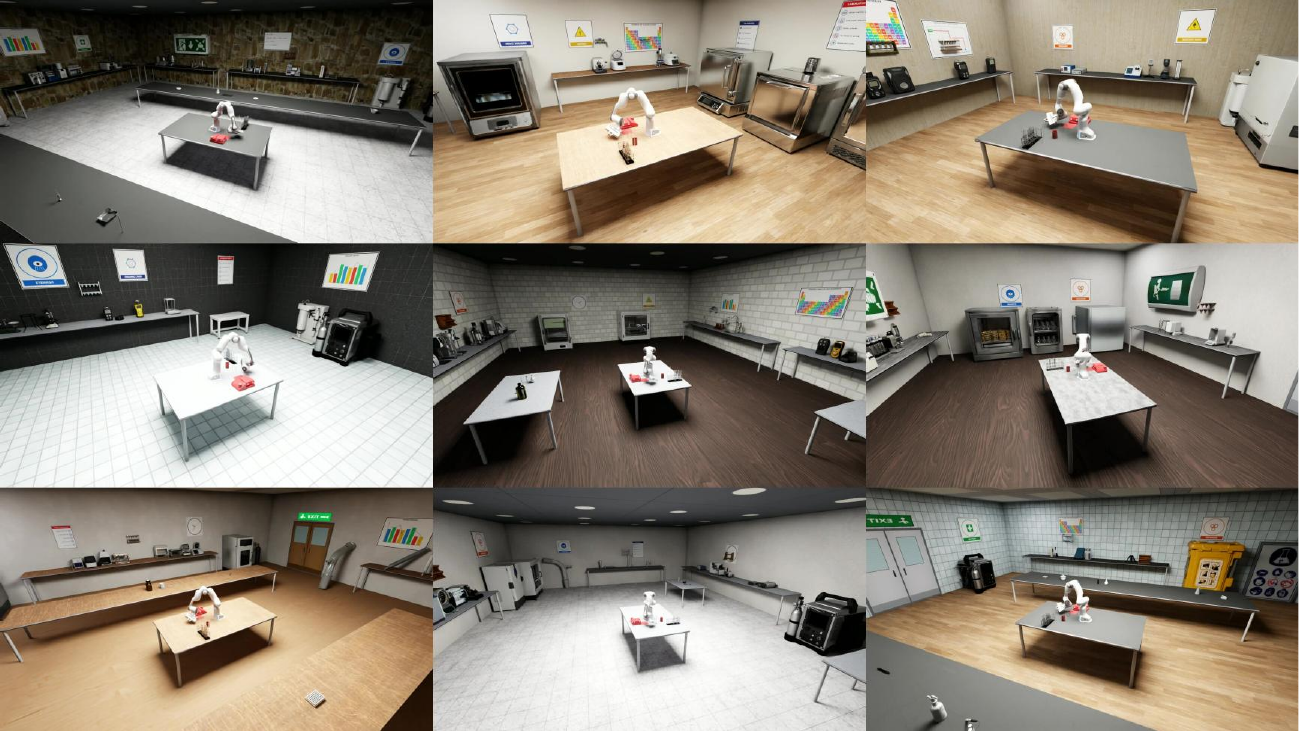}
    \caption{Nine laboratory scenes generated by RoboGenesis with domain randomization.
    Scenes vary in room layout, bench geometry, flooring material, wall texture, lighting condition, instrument placement, and background objects such as posters, cabinets, and safety signs.
    Each scene hosts a single arm robot with randomized labware on the bench.
    This diversity reduces visual overfitting so that the trained policy generalizes across laboratory environments rather than memorizing a fixed scene configuration.}
    \label{fig:scene-diversity}
\end{figure*}

\subsection{RoboGenesis Scene Diversity}\label{app:scene-diversity}

\Cref{fig:scene-diversity} visualizes nine laboratory scenes produced by RoboGenesis under domain randomization.
Each scene is constructed by sampling room geometry, bench layout, flooring and wall materials, lighting conditions, and the placement and type of laboratory instruments and background objects.
This randomization targets the visual factors that differ most between real laboratories: bench surfaces, overhead lighting color temperature, instrument clutter, and wall mounted items such as posters and safety signs.
This diversity drives the OOD generalization reported in \Cref{tab:level3_results}: the policy sees a wide distribution of visual contexts during training and is less likely to overfit to a single scene layout or texture palette.

\section{Asset Generation Prompt Template}\label{app:asset-prompt}

The text to image prompt used by the RoboGenesis asset generation pipeline (\Cref{sec:robogenesis}) follows the template below.
Placeholders in angle brackets are filled from the text description of each target object: \texttt{item} is the object name with optional size and color qualifiers, \texttt{features} lists 3--6 distinguishing visual attributes, \texttt{material} expands to a physically grounded material description (e.g.\ ``crystal clear borosilicate glass with subtle specular highlights and visible rim thickness''), and \texttt{viewpoint} is sampled from five three quarter angle variants to diversify reconstruction input.

\begin{tcolorbox}[
  breakable,
  colback=primary!3,
  colframe=primary!70,
  coltitle=white,
  fonttitle=\bfseries\small,
  title=Text to Image Prompt Template for Asset Generation,
  boxrule=0.6pt,
  arc=2pt,
  left=6pt, right=6pt, top=4pt, bottom=4pt,
]
\small\ttfamily
A professional product photograph of a single \textit{\textrm{<item>}}, \textit{\textrm{<features>}}, \textit{\textrm{<material>}}, soft diffused studio lighting, isolated on a pure clean white seamless background, \textit{\textrm{<viewpoint>}}, hyperrealistic, photorealistic product photography, sharp focus, 8K detail, no text, no watermark, no label, no background objects, no multiple items, suitable for 3D reconstruction
\end{tcolorbox}

\section{Scene Construction Placement Rules}\label{app:placement-rules}

\Cref{tab:placement-rules} lists the numerical constraints that the RoboGenesis scene solver enforces when placing assets.
Values are read from the constraint solver source code; the validation pass computes a 0--100 quality score (fraction of checks passed) and rejects scenes below threshold.

\begin{table}[h]
\centering
\footnotesize
\caption{Asset placement rules and validation constraints used by the RoboGenesis scene construction solver.}\label{tab:placement-rules}
\renewcommand{\arraystretch}{1.08}
\begin{tabular}{p{0.26\linewidth} p{0.36\linewidth} p{0.22\linewidth}}
\toprule
\textbf{Category} & \textbf{Rule} & \textbf{Constraint} \\
\midrule
\multicolumn{3}{l}{\textit{Room layout}} \\
Small room & Width $\times$ depth & 7.5--10.0\,m $\times$ 6.0--8.0\,m \\
Large room & Width $\times$ depth & 11.0--15.0\,m $\times$ 8.5--11.0\,m \\
Ceiling height & Floor to ceiling & 2.8--3.5\,m \\
\midrule
\multicolumn{3}{l}{\textit{Central work tables}} \\
Table height & Work surface & 0.71\,m \\
Intertable gap & Minimum clearance & $\geq$ 0.8\,m \\
Perimeter margin & Wall furniture setback & 1.5\,m \\
\midrule
\multicolumn{3}{l}{\textit{Wall counters}} \\
Counter height & Surface height & 0.9\,m \\
Counter depth & Front to wall & 0.65\,m \\
Counter--wall gap & Back edge to wall & 0.06\,m \\
Corner margin & Keep corners clear & 0.8\,m \\
\midrule
\multicolumn{3}{l}{\textit{Counter equipment (clusters)}} \\
Cluster size & Items per cluster & 2--4 \\
Intracluster gap & Between items & 4--12\,cm \\
Intercluster gap & Between clusters & 50--100\,cm \\
\midrule
\multicolumn{3}{l}{\textit{Floor furniture}} \\
Wall gap & Back edge to wall & 0.12\,m \\
Robot aisle & Clearance to central tables & $\geq$ 0.6\,m \\
\midrule
\multicolumn{3}{l}{\textit{Wall mounted items}} \\
Mount height & Bottom of item & 1.5\,m \\
\midrule
\multicolumn{3}{l}{\textit{Glassware on tables}} \\
Max item dimension & Per object footprint & $\leq$ 0.4\,m \\
Edge clear band & Large room tables & 0.3\,m \\
\midrule
\multicolumn{3}{l}{\textit{Door}} \\
Single leaf width & Standard scenes & 1.1\,m \\
Double leaf width & Navigation scenes & 1.8\,m \\
Clearance depth & Open space in front & 1.1\,m \\
Side margin & Clear span each side & 0.35\,m \\
\midrule
\multicolumn{3}{l}{\textit{Validation checks (pass/fail, scored 0--100)}} \\
Work table gap & Pairwise $\geq$ 0.8\,m & pass/fail \\
Robot aisle & Perimeter--table $\geq$ 0.5\,m & pass/fail \\
Robot placement & Robot on main table & pass/fail \\
Counter presence & $\geq$ 1 counter & pass/fail \\
Floor asset size & All max\_dim $\geq$ 0.5\,m & pass/fail \\
Floor asset count & $\geq$ 4 items & pass/fail \\
Table grounding & Items resting on surface & pass/fail \\
Floor grounding & Bottom $\approx$ Z\,=\,0 & pass/fail \\
No floor overlap & Pairwise collision free & pass/fail \\
No wall clip & All inside room bounds & pass/fail \\
\bottomrule
\end{tabular}
\end{table}

%% file: labvla.bbl
\begin{thebibliography}{85}
\providecommand{\natexlab}[1]{#1}
\providecommand{\url}[1]{\texttt{#1}}
\expandafter\ifx\csname urlstyle\endcsname\relax
  \providecommand{\doi}[1]{doi: #1}\else
  \providecommand{\doi}{doi: \begingroup \urlstyle{rm}\Url}\fi

\bibitem[Bai et~al.(2025)Bai, Cai, Chen, Chen, Chen, Cheng, Deng, Ding, Gao, Ge, et~al.]{bai2025qwen3}
Shuai Bai, Yuxuan Cai, Ruizhe Chen, Keqin Chen, Xionghui Chen, Zesen Cheng, Lianghao Deng, Wei Ding, Chang Gao, Chunjiang Ge, et~al.
\newblock Qwen3-vl technical report.
\newblock \emph{arXiv preprint arXiv:2511.21631}, 2025.

\bibitem[Bjorck et~al.(2025)Bjorck, Casta{\~n}eda, Cherniadev, Da, Ding, Fan, Fang, Fox, Hu, Huang, et~al.]{bjorck2025gr00t}
Johan Bjorck, Fernando Casta{\~n}eda, Nikita Cherniadev, Xingye Da, Runyu Ding, Linxi Fan, Yu~Fang, Dieter Fox, Fengyuan Hu, Spencer Huang, et~al.
\newblock Gr00t n1: An open foundation model for generalist humanoid robots.
\newblock \emph{arXiv preprint arXiv:2503.14734}, 2025.

\bibitem[Black et~al.(2024)Black, Brown, Driess, Esmail, Equi, Finn, Fusai, Groom, Hausman, Ichter, et~al.]{black2024pi_0}
Kevin Black, Noah Brown, Danny Driess, Adnan Esmail, Michael Equi, Chelsea Finn, Niccolo Fusai, Lachy Groom, Karol Hausman, Brian Ichter, et~al.
\newblock $\pi_0$: A vision-language-action flow model for general robot control.
\newblock \emph{arXiv preprint arXiv:2410.24164}, 2024.

\bibitem[Boiko et~al.(2023)Boiko, MacKnight, Kline, and Gomes]{boiko2023autonomous}
Daniil~A Boiko, Robert MacKnight, Ben Kline, and Gabe Gomes.
\newblock Autonomous chemical research with large language models.
\newblock \emph{Nature}, 624\penalty0 (7992):\penalty0 570--578, 2023.

\bibitem[Brohan et~al.(2022)Brohan, Brown, Carbajal, Chebotar, Dabis, Finn, Gopalakrishnan, Hausman, Herzog, Hsu, et~al.]{brohan2022rt}
Anthony Brohan, Noah Brown, Justice Carbajal, Yevgen Chebotar, Joseph Dabis, Chelsea Finn, Keerthana Gopalakrishnan, Karol Hausman, Alex Herzog, Jasmine Hsu, et~al.
\newblock Rt-1: Robotics transformer for real-world control at scale.
\newblock \emph{arXiv preprint arXiv:2212.06817}, 2022.

\bibitem[Bu et~al.(2025{\natexlab{a}})Bu, Cai, Chen, Cui, Ding, Feng, Gao, He, Hu, Huang, et~al.]{bu2025agibot}
Qingwen Bu, Jisong Cai, Li~Chen, Xiuqi Cui, Yan Ding, Siyuan Feng, Shenyuan Gao, Xindong He, Xuan Hu, Xu~Huang, et~al.
\newblock Agibot world colosseo: A large-scale manipulation platform for scalable and intelligent embodied systems.
\newblock \emph{arXiv preprint arXiv:2503.06669}, 2025{\natexlab{a}}.

\bibitem[Bu et~al.(2025{\natexlab{b}})Bu, Yang, Cai, Gao, Ren, Yao, Luo, and Li]{bu2025univla}
Qingwen Bu, Yanting Yang, Jisong Cai, Shenyuan Gao, Guanghui Ren, Maoqing Yao, Ping Luo, and Hongyang Li.
\newblock Univla: Learning to act anywhere with task-centric latent actions.
\newblock \emph{arXiv preprint arXiv:2505.06111}, 2025{\natexlab{b}}.

\bibitem[Burger et~al.(2020)Burger, Maffettone, Gusev, Aitchison, Bai, Wang, Li, Alston, Li, Clowes, et~al.]{burger2020mobile}
Benjamin Burger, Phillip~M Maffettone, Vladimir~V Gusev, Catherine~M Aitchison, Yang Bai, Xiaoyan Wang, Xiaobo Li, Ben~M Alston, Buyi Li, Rob Clowes, et~al.
\newblock A mobile robotic chemist.
\newblock \emph{Nature}, 583\penalty0 (7815):\penalty0 237--241, 2020.

\bibitem[Cai et~al.(2026)Cai, Cai, Cao, Chen, He, Jiang, Li, Li, Li, Liu, et~al.]{cai2026internvla}
Junhao Cai, Zetao Cai, Jiafei Cao, Yilun Chen, Zeyu He, Lei Jiang, Hang Li, Hengjie Li, Yang Li, Yufei Liu, et~al.
\newblock Internvla-a1: Unifying understanding, generation and action for robotic manipulation.
\newblock \emph{arXiv preprint arXiv:2601.02456}, 2026.

\bibitem[Chen et~al.(2025)Chen, Chen, Chen, Cai, Liu, Li, Liang, Lin, Ge, Gu, et~al.]{chen2025robotwin}
Tianxing Chen, Zanxin Chen, Baijun Chen, Zijian Cai, Yibin Liu, Zixuan Li, Qiwei Liang, Xianliang Lin, Yiheng Ge, Zhenyu Gu, et~al.
\newblock Robotwin 2.0: A scalable data generator and benchmark with strong domain randomization for robust bimanual robotic manipulation.
\newblock \emph{arXiv preprint arXiv:2506.18088}, 2025.

\bibitem[Dasari et~al.(2019)Dasari, Ebert, Tian, Nair, Bucher, Schmeckpeper, Singh, Levine, and Finn]{dasari2019robonet}
Sudeep Dasari, Frederik Ebert, Stephen Tian, Suraj Nair, Bernadette Bucher, Karl Schmeckpeper, Siddharth Singh, Sergey Levine, and Chelsea Finn.
\newblock Robonet: Large-scale multi-robot learning.
\newblock \emph{arXiv preprint arXiv:1910.11215}, 2019.

\bibitem[Driess et~al.(2026)Driess, Springenberg, Ichter, Yu, Li-Bell, Pertsch, Ren, Walke, Vuong, Shi, et~al.]{driess2026knowledge}
Danny Driess, Jost Springenberg, Brian Ichter, Lili Yu, Adrian Li-Bell, Karl Pertsch, Allen Ren, Homer Walke, Quan Vuong, Lucy~Xiaoyang Shi, et~al.
\newblock Knowledge insulating vision-language-action models: Train fast, run fast, generalize better.
\newblock \emph{Advances in Neural Information Processing Systems}, 38:\penalty0 102867--102888, 2026.

\bibitem[Gervet et~al.(2023)Gervet, Xian, Gkanatsios, and Fragkiadaki]{gervet2023act3d}
Theophile Gervet, Zhou Xian, Nikolaos Gkanatsios, and Katerina Fragkiadaki.
\newblock Act3d: 3d feature field transformers for multi-task robotic manipulation.
\newblock \emph{arXiv preprint arXiv:2306.17817}, 2023.

\bibitem[Goyal et~al.(2023)Goyal, Xu, Guo, Blukis, Chao, and Fox]{goyal2023rvt}
Ankit Goyal, Jie Xu, Yijie Guo, Valts Blukis, Yu-Wei Chao, and Dieter Fox.
\newblock Rvt: Robotic view transformer for 3d object manipulation.
\newblock In \emph{Conference on Robot Learning}, pages 694--710. PMLR, 2023.

\bibitem[Goyal et~al.(2024)Goyal, Blukis, Xu, Guo, Chao, and Fox]{goyal2024rvt}
Ankit Goyal, Valts Blukis, Jie Xu, Yijie Guo, Yu-Wei Chao, and Dieter Fox.
\newblock Rvt-2: Learning precise manipulation from few demonstrations.
\newblock \emph{arXiv preprint arXiv:2406.08545}, 2024.

\bibitem[Granda et~al.(2018)Granda, Donina, Dragone, Long, and Cronin]{granda2018controlling}
Jaros{\l}aw~M Granda, Liva Donina, Vincenza Dragone, De-Liang Long, and Leroy Cronin.
\newblock Controlling an organic synthesis robot with machine learning to search for new reactivity.
\newblock \emph{Nature}, 559\penalty0 (7714):\penalty0 377--381, 2018.

\bibitem[Gu et~al.(2023)Gu, Xiang, Li, Ling, Liu, Mu, Tang, Tao, Wei, Yao, et~al.]{gu2023maniskill2}
Jiayuan Gu, Fanbo Xiang, Xuanlin Li, Zhan Ling, Xiqiang Liu, Tongzhou Mu, Yihe Tang, Stone Tao, Xinyue Wei, Yunchao Yao, et~al.
\newblock Maniskill2: A unified benchmark for generalizable manipulation skills.
\newblock \emph{arXiv preprint arXiv:2302.04659}, 2023.

\bibitem[Heo et~al.(2025)Heo, Lee, Lee, and Lim]{heo2025furniturebench}
Minho Heo, Youngwoon Lee, Doohyun Lee, and Joseph~J Lim.
\newblock Furniturebench: Reproducible real-world benchmark for long-horizon complex manipulation.
\newblock \emph{The International Journal of Robotics Research}, 44\penalty0 (10-11):\penalty0 1863--1891, 2025.

\bibitem[Hsu et~al.(2024)Hsu, Dai, Kothapalli, Song, Tang, Zhu, Shimizu, Sahni, Ning, and Chen]{hsu2024liger}
Pin-Lun Hsu, Yun Dai, Vignesh Kothapalli, Qingquan Song, Shao Tang, Siyu Zhu, Steven Shimizu, Shivam Sahni, Haowen Ning, and Yanning Chen.
\newblock Liger kernel: Efficient triton kernels for llm training.
\newblock \emph{arXiv preprint arXiv:2410.10989}, 2024.

\bibitem[Huang et~al.(2026)Huang, Wu, Chen, Wang, and Yang]{huang2026thinkact}
Chi-Pin Huang, Yueh-Hua Wu, Min-Hung Chen, Frank Wang, and Fu-En Yang.
\newblock Thinkact: Vision-language-action reasoning via reinforced visual latent planning.
\newblock \emph{Advances in Neural Information Processing Systems}, 38:\penalty0 82782--82802, 2026.

\bibitem[Huang et~al.(2025)Huang, Pipe, Martin, Wang, Franklin, Tyrrell, Fairlamb, and Zhu]{huang2025tarmac}
Kefeng Huang, Jonathon Pipe, Alice~E Martin, Tianyuan Wang, Barnabas~A Franklin, Andy~M Tyrrell, Ian~JS Fairlamb, and Jihong Zhu.
\newblock Tarmac: A taxonomy for robot manipulation in chemistry.
\newblock \emph{arXiv preprint arXiv:2510.19289}, 2025.

\bibitem[Intelligence et~al.(2025)Intelligence, Black, Brown, Darpinian, Dhabalia, Driess, Esmail, Equi, Finn, Fusai, et~al.]{intelligence2025pi_}
Physical Intelligence, Kevin Black, Noah Brown, James Darpinian, Karan Dhabalia, Danny Driess, Adnan Esmail, Michael Equi, Chelsea Finn, Niccolo Fusai, et~al.
\newblock ${\pi}_{0.5}$: a vision-language-action model with open-world generalization.
\newblock \emph{arXiv preprint arXiv:2504.16054}, 2025.

\bibitem[James et~al.(2020)James, Ma, Arrojo, and Davison]{james2020rlbench}
Stephen James, Zicong Ma, David~Rovick Arrojo, and Andrew~J Davison.
\newblock Rlbench: The robot learning benchmark \& learning environment.
\newblock \emph{IEEE Robotics and Automation Letters}, 5\penalty0 (2):\penalty0 3019--3026, 2020.

\bibitem[Jang et~al.(2022)Jang, Irpan, Khansari, Kappler, Ebert, Lynch, Levine, and Finn]{jang2022bc}
Eric Jang, Alex Irpan, Mohi Khansari, Daniel Kappler, Frederik Ebert, Corey Lynch, Sergey Levine, and Chelsea Finn.
\newblock Bc-z: Zero-shot task generalization with robotic imitation learning.
\newblock In \emph{conference on Robot Learning}, pages 991--1002. PMLR, 2022.

\bibitem[Jiang et~al.(2023)Jiang, Gupta, Zhang, Wang, Dou, Chen, Fei-Fei, Anandkumar, Zhu, and Fan]{jiang2023vima}
Yunfan Jiang, Agrim Gupta, Zichen Zhang, Guanzhi Wang, Yongqiang Dou, Yanjun Chen, Li~Fei-Fei, Anima Anandkumar, Yuke Zhu, and Linxi Fan.
\newblock Vima: Robot manipulation with multimodal prompts.
\newblock 2023.

\bibitem[Jumper et~al.(2021)Jumper, Evans, Pritzel, Green, Figurnov, Ronneberger, Tunyasuvunakool, Bates, {\v{Z}}{\'\i}dek, Potapenko, et~al.]{jumper2021highly}
John Jumper, Richard Evans, Alexander Pritzel, Tim Green, Michael Figurnov, Olaf Ronneberger, Kathryn Tunyasuvunakool, Russ Bates, Augustin {\v{Z}}{\'\i}dek, Anna Potapenko, et~al.
\newblock Highly accurate protein structure prediction with alphafold.
\newblock \emph{nature}, 596\penalty0 (7873):\penalty0 583--589, 2021.

\bibitem[Khazatsky et~al.(2024)Khazatsky, Pertsch, Nair, Balakrishna, Dasari, Karamcheti, Nasiriany, Srirama, Chen, Ellis, et~al.]{khazatsky2024droid}
Alexander Khazatsky, Karl Pertsch, Suraj Nair, Ashwin Balakrishna, Sudeep Dasari, Siddharth Karamcheti, Soroush Nasiriany, Mohan~Kumar Srirama, Lawrence~Yunliang Chen, Kirsty Ellis, et~al.
\newblock Droid: A large-scale in-the-wild robot manipulation dataset.
\newblock \emph{arXiv preprint arXiv:2403.12945}, 2024.

\bibitem[Kim et~al.(2024)Kim, Pertsch, Karamcheti, Xiao, Balakrishna, Nair, Rafailov, Foster, Lam, Sanketi, et~al.]{kim2024openvla}
Moo~Jin Kim, Karl Pertsch, Siddharth Karamcheti, Ted Xiao, Ashwin Balakrishna, Suraj Nair, Rafael Rafailov, Ethan Foster, Grace Lam, Pannag Sanketi, et~al.
\newblock Openvla: An open-source vision-language-action model.
\newblock \emph{arXiv preprint arXiv:2406.09246}, 2024.

\bibitem[Kim et~al.(2025)Kim, Finn, and Liang]{kim2025fine}
Moo~Jin Kim, Chelsea Finn, and Percy Liang.
\newblock Fine-tuning vision-language-action models: Optimizing speed and success.
\newblock \emph{arXiv preprint arXiv:2502.19645}, 2025.

\bibitem[Li et~al.(2021)Li, Xia, Mart{\'\i}n-Mart{\'\i}n, Lingelbach, Srivastava, Shen, Vainio, Gokmen, Dharan, Jain, et~al.]{li2021igibson}
Chengshu Li, Fei Xia, Roberto Mart{\'\i}n-Mart{\'\i}n, Michael Lingelbach, Sanjana Srivastava, Bokui Shen, Kent Vainio, Cem Gokmen, Gokul Dharan, Tanish Jain, et~al.
\newblock igibson 2.0: Object-centric simulation for robot learning of everyday household tasks.
\newblock \emph{arXiv preprint arXiv:2108.03272}, 2021.

\bibitem[Li et~al.(2023)Li, Zhang, Wong, Gokmen, Srivastava, Mart{\'\i}n-Mart{\'\i}n, Wang, Levine, Lingelbach, Sun, et~al.]{li2023behavior}
Chengshu Li, Ruohan Zhang, Josiah Wong, Cem Gokmen, Sanjana Srivastava, Roberto Mart{\'\i}n-Mart{\'\i}n, Chen Wang, Gabrael Levine, Michael Lingelbach, Jiankai Sun, et~al.
\newblock Behavior-1k: A benchmark for embodied ai with 1,000 everyday activities and realistic simulation.
\newblock In \emph{Conference on Robot Learning}, pages 80--93. PMLR, 2023.

\bibitem[Li et~al.(2025{\natexlab{a}})Li, Zhu, Tang, Wen, Zhu, Liu, Li, Cheng, Peng, Peng, et~al.]{li2025coa}
Jinming Li, Yichen Zhu, Zhibin Tang, Junjie Wen, Minjie Zhu, Xiaoyu Liu, Chengmeng Li, Ran Cheng, Yaxin Peng, Yan Peng, et~al.
\newblock Coa-vla: Improving vision-language-action models via visual-text chain-of-affordance.
\newblock In \emph{Proceedings of the IEEE/CVF International Conference on Computer Vision}, pages 9759--9769, 2025{\natexlab{a}}.

\bibitem[Li et~al.(2024{\natexlab{a}})Li, Liang, Wang, Luo, Chen, Liao, Wei, Deng, Xu, Zhang, et~al.]{li2024cogact}
Qixiu Li, Yaobo Liang, Zeyu Wang, Lin Luo, Xi~Chen, Mozheng Liao, Fangyun Wei, Yu~Deng, Sicheng Xu, Yizhong Zhang, et~al.
\newblock Cogact: A foundational vision-language-action model for synergizing cognition and action in robotic manipulation.
\newblock \emph{arXiv preprint arXiv:2411.19650}, 2024{\natexlab{a}}.

\bibitem[Li et~al.(2026)Li, Hu, Qu, Zhang, Yin, Zhang, Huang, Wang, Wang, Pang, et~al.]{li2026labutopia}
Rui Li, Zixuan Hu, Wenxi Qu, Jinouwen Zhang, Zhenfei Yin, Sha Zhang, Xuantuo Huang, Hanqing Wang, Tai Wang, Jiangmiao Pang, et~al.
\newblock Labutopia: High-fidelity simulation and hierarchical benchmark for scientific embodied agents.
\newblock \emph{Advances in Neural Information Processing Systems}, 38, 2026.

\bibitem[Li et~al.(2025{\natexlab{b}})Li, Guo, Wu, Wang, Deng, Weng, Tan, and Wang]{li2025map}
Runhao Li, Wenkai Guo, Zhenyu Wu, Changyuan Wang, Haoyuan Deng, Zhenyu Weng, Yap-Peng Tan, and Ziwei Wang.
\newblock Map-vla: Memory-augmented prompting for vision-language-action model in robotic manipulation.
\newblock \emph{arXiv preprint arXiv:2511.09516}, 2025{\natexlab{b}}.

\bibitem[Li et~al.(2024{\natexlab{b}})Li, Huang, Guo, Wu, Zhang, Zhang, and Ding]{li2024chemistry3d}
Shoujie Li, Yan Huang, Changqing Guo, Tong Wu, Jiawei Zhang, Linrui Zhang, and Wenbo Ding.
\newblock Chemistry3d: Robotic interaction benchmark for chemistry experiments.
\newblock \emph{arXiv preprint arXiv:2406.08160}, 2024{\natexlab{b}}.

\bibitem[Liang et~al.(2025)Liang, Sun, He, Dong, Dai, Laptev, Khan, and Cong]{liang2025pixelvla}
Wenqi Liang, Gan Sun, Yao He, Jiahua Dong, Suyan Dai, Ivan Laptev, Salman Khan, and Yang Cong.
\newblock Pixelvla: Advancing pixel-level understanding in vision-language-action model.
\newblock \emph{arXiv preprint arXiv:2511.01571}, 2025.

\bibitem[Lin et~al.(2021)Lin, Wang, Olkin, and Held]{lin2021softgym}
Xingyu Lin, Yufei Wang, Jake Olkin, and David Held.
\newblock Softgym: Benchmarking deep reinforcement learning for deformable object manipulation.
\newblock In \emph{Conference on Robot Learning}, pages 432--448. PMLR, 2021.

\bibitem[Lipman et~al.(2022)Lipman, Chen, Ben-Hamu, Nickel, and Le]{lipman2022flow}
Yaron Lipman, Ricky~TQ Chen, Heli Ben-Hamu, Maximilian Nickel, and Matt Le.
\newblock Flow matching for generative modeling.
\newblock \emph{arXiv preprint arXiv:2210.02747}, 2022.

\bibitem[Liu et~al.(2023)Liu, Zhu, Gao, Feng, Liu, Zhu, and Stone]{liu2023libero}
Bo~Liu, Yifeng Zhu, Chongkai Gao, Yihao Feng, Qiang Liu, Yuke Zhu, and Peter Stone.
\newblock Libero: Benchmarking knowledge transfer for lifelong robot learning.
\newblock \emph{Advances in Neural Information Processing Systems}, 36:\penalty0 44776--44791, 2023.

\bibitem[Liu et~al.(2024)Liu, Liu, Wang, An, Li, Zhou, Yang, Zhang, Guo, and Zhang]{liu2024robomamba}
Jiaming Liu, Mengzhen Liu, Zhenyu Wang, Pengju An, Xiaoqi Li, Kaichen Zhou, Senqiao Yang, Renrui Zhang, Yandong Guo, and Shanghang Zhang.
\newblock Robomamba: Efficient vision-language-action model for robotic reasoning and manipulation.
\newblock \emph{Advances in Neural Information Processing Systems}, 37:\penalty0 40085--40110, 2024.

\bibitem[Liu et~al.(2025)Liu, Wu, Li, Tan, Chen, Wang, Xu, Su, and Zhu]{liu2025rdt}
Songming Liu, Lingxuan Wu, Bangguo Li, Hengkai Tan, Huayu Chen, Zhengyi Wang, Ke~Xu, Hang Su, and Jun Zhu.
\newblock Rdt-1b: a diffusion foundation model for bimanual manipulation.
\newblock In \emph{International Conference on Learning Representations}, volume 2025, pages 29982--30009, 2025.

\bibitem[Liu et~al.(2022)Liu, Gong, and Liu]{liu2022flow}
Xingchao Liu, Chengyue Gong, and Qiang Liu.
\newblock Flow straight and fast: Learning to generate and transfer data with rectified flow.
\newblock \emph{arXiv preprint arXiv:2209.03003}, 2022.

\bibitem[Lu et~al.(2024)Lu, Lu, Lange, Foerster, Clune, and Ha]{lu2024ai}
Chris Lu, Cong Lu, Robert~Tjarko Lange, Jakob Foerster, Jeff Clune, and David Ha.
\newblock The ai scientist: Towards fully automated open-ended scientific discovery.
\newblock \emph{arXiv preprint arXiv:2408.06292}, 2024.

\bibitem[Luo et~al.(2025)Luo, Xu, Liu, Tan, Lin, Wu, Abbeel, and Levine]{luo2025fmb}
Jianlan Luo, Charles Xu, Fangchen Liu, Liam Tan, Zipeng Lin, Jeffrey Wu, Pieter Abbeel, and Sergey Levine.
\newblock Fmb: a functional manipulation benchmark for generalizable robotic learning.
\newblock \emph{The International Journal of Robotics Research}, 44\penalty0 (4):\penalty0 592--606, 2025.

\bibitem[M.~Bran et~al.(2024)M.~Bran, Cox, Schilter, Baldassari, White, and Schwaller]{m2024augmenting}
Andres M.~Bran, Sam Cox, Oliver Schilter, Carlo Baldassari, Andrew~D White, and Philippe Schwaller.
\newblock Augmenting large language models with chemistry tools.
\newblock \emph{Nature machine intelligence}, 6\penalty0 (5):\penalty0 525--535, 2024.

\bibitem[Mees et~al.(2022)Mees, Hermann, Rosete-Beas, and Burgard]{mees2022calvin}
Oier Mees, Lukas Hermann, Erick Rosete-Beas, and Wolfram Burgard.
\newblock Calvin: A benchmark for language-conditioned policy learning for long-horizon robot manipulation tasks.
\newblock \emph{IEEE Robotics and Automation Letters}, 7\penalty0 (3):\penalty0 7327--7334, 2022.

\bibitem[Mehr et~al.(2020)Mehr, Craven, Leonov, Keenan, and Cronin]{mehr2020universal}
S~Hessam~M Mehr, Matthew Craven, Artem~I Leonov, Graham Keenan, and Leroy Cronin.
\newblock A universal system for digitization and automatic execution of the chemical synthesis literature.
\newblock \emph{Science}, 370\penalty0 (6512):\penalty0 101--108, 2020.

\bibitem[Merchant et~al.(2023)Merchant, Batzner, Schoenholz, Aykol, Cheon, and Cubuk]{merchant2023scaling}
Amil Merchant, Simon Batzner, Samuel~S Schoenholz, Muratahan Aykol, Gowoon Cheon, and Ekin~Dogus Cubuk.
\newblock Scaling deep learning for materials discovery.
\newblock \emph{Nature}, 624\penalty0 (7990):\penalty0 80--85, 2023.

\bibitem[Nasiriany et~al.(2024)Nasiriany, Maddukuri, Zhang, Parikh, Lo, Joshi, Mandlekar, and Zhu]{nasiriany2024robocasa}
Soroush Nasiriany, Abhiram Maddukuri, Lance Zhang, Adeet Parikh, Aaron Lo, Abhishek Joshi, Ajay Mandlekar, and Yuke Zhu.
\newblock Robocasa: Large-scale simulation of everyday tasks for generalist robots.
\newblock \emph{arXiv preprint arXiv:2406.02523}, 2024.

\bibitem[Nasiriany et~al.(2026)Nasiriany, Nasiriany, Maddukuri, and Zhu]{nasiriany2026robocasa365}
Soroush Nasiriany, Sepehr Nasiriany, Abhiram Maddukuri, and Yuke Zhu.
\newblock Robocasa365: A large-scale simulation framework for training and benchmarking generalist robots.
\newblock \emph{arXiv preprint arXiv:2603.04356}, 2026.

\bibitem[O’Neill et~al.(2024)O’Neill, Rehman, Maddukuri, Gupta, Padalkar, Lee, Pooley, Gupta, Mandlekar, Jain, et~al.]{o2024open}
Abby O’Neill, Abdul Rehman, Abhiram Maddukuri, Abhishek Gupta, Abhishek Padalkar, Abraham Lee, Acorn Pooley, Agrim Gupta, Ajay Mandlekar, Ajinkya Jain, et~al.
\newblock Open x-embodiment: Robotic learning datasets and rt-x models: Open x-embodiment collaboration 0.
\newblock In \emph{2024 IEEE International Conference on Robotics and Automation (ICRA)}, pages 6892--6903. IEEE, 2024.

\bibitem[Peebles and Xie(2023)]{peebles2023scalable}
William Peebles and Saining Xie.
\newblock Scalable diffusion models with transformers.
\newblock In \emph{Proceedings of the IEEE/CVF international conference on computer vision}, pages 4195--4205, 2023.

\bibitem[Pertsch et~al.(2025)Pertsch, Stachowicz, Ichter, Driess, Nair, Vuong, Mees, Finn, and Levine]{pertsch2025fast}
Karl Pertsch, Kyle Stachowicz, Brian Ichter, Danny Driess, Suraj Nair, Quan Vuong, Oier Mees, Chelsea Finn, and Sergey Levine.
\newblock Fast: Efficient action tokenization for vision-language-action models.
\newblock \emph{arXiv preprint arXiv:2501.09747}, 2025.

\bibitem[Qu et~al.(2025)Qu, Song, Chen, Yao, Ye, Ding, Wang, Gu, Zhao, Wang, et~al.]{qu2025spatialvla}
Delin Qu, Haoming Song, Qizhi Chen, Yuanqi Yao, Xinyi Ye, Yan Ding, Zhigang Wang, JiaYuan Gu, Bin Zhao, Dong Wang, et~al.
\newblock Spatialvla: Exploring spatial representations for visual-language-action model.
\newblock \emph{arXiv preprint arXiv:2501.15830}, 2025.

\bibitem[Shi et~al.(2025)Shi, Xie, Liu, Sun, Liu, Wang, Zhou, Fan, Zhang, and Huang]{shi2025memoryvla}
Hao Shi, Bin Xie, Yingfei Liu, Lin Sun, Fengrong Liu, Tiancai Wang, Erjin Zhou, Haoqiang Fan, Xiangyu Zhang, and Gao Huang.
\newblock Memoryvla: Perceptual-cognitive memory in vision-language-action models for robotic manipulation.
\newblock \emph{arXiv preprint arXiv:2508.19236}, 2025.

\bibitem[Shridhar et~al.(2020)Shridhar, Thomason, Gordon, Bisk, Han, Mottaghi, Zettlemoyer, and Fox]{shridhar2020alfred}
Mohit Shridhar, Jesse Thomason, Daniel Gordon, Yonatan Bisk, Winson Han, Roozbeh Mottaghi, Luke Zettlemoyer, and Dieter Fox.
\newblock Alfred: A benchmark for interpreting grounded instructions for everyday tasks.
\newblock In \emph{Proceedings of the IEEE/CVF conference on computer vision and pattern recognition}, pages 10740--10749, 2020.

\bibitem[Shridhar et~al.(2022)Shridhar, Manuelli, and Fox]{shridhar2022cliport}
Mohit Shridhar, Lucas Manuelli, and Dieter Fox.
\newblock Cliport: What and where pathways for robotic manipulation.
\newblock In \emph{Conference on robot learning}, pages 894--906. PMLR, 2022.

\bibitem[Shridhar et~al.(2023)Shridhar, Manuelli, and Fox]{shridhar2023perceiver}
Mohit Shridhar, Lucas Manuelli, and Dieter Fox.
\newblock Perceiver-actor: A multi-task transformer for robotic manipulation.
\newblock In \emph{Conference on Robot Learning}, pages 785--799. PMLR, 2023.

\bibitem[Shukor et~al.(2025)Shukor, Aubakirova, Capuano, Kooijmans, Palma, Zouitine, Aractingi, Pascal, Russi, Marafioti, et~al.]{shukor2025smolvla}
Mustafa Shukor, Dana Aubakirova, Francesco Capuano, Pepijn Kooijmans, Steven Palma, Adil Zouitine, Michel Aractingi, Caroline Pascal, Martino Russi, Andres Marafioti, et~al.
\newblock Smolvla: A vision-language-action model for affordable and efficient robotics.
\newblock \emph{arXiv preprint arXiv:2506.01844}, 2025.

\bibitem[Szot et~al.(2021)Szot, Clegg, Undersander, Wijmans, Zhao, Turner, Maestre, Mukadam, Chaplot, Maksymets, et~al.]{szot2021habitat}
Andrew Szot, Alexander Clegg, Eric Undersander, Erik Wijmans, Yili Zhao, John Turner, Noah Maestre, Mustafa Mukadam, Devendra~Singh Chaplot, Oleksandr Maksymets, et~al.
\newblock Habitat 2.0: Training home assistants to rearrange their habitat.
\newblock \emph{Advances in neural information processing systems}, 34:\penalty0 251--266, 2021.

\bibitem[Szymanski et~al.(2023)Szymanski, Rendy, Fei, Kumar, He, Milsted, McDermott, Gallant, Cubuk, Merchant, et~al.]{szymanski2023autonomous}
Nathan~J Szymanski, Bernardus Rendy, Yuxing Fei, Rishi~E Kumar, Tanjin He, David Milsted, Matthew~J McDermott, Max Gallant, Ekin~Dogus Cubuk, Amil Merchant, et~al.
\newblock An autonomous laboratory for the accelerated synthesis of inorganic materials.
\newblock \emph{Nature}, 624\penalty0 (7990):\penalty0 86, 2023.

\bibitem[Tao et~al.(2024)Tao, Xiang, Shukla, Qin, Hinrichsen, Yuan, Bao, Lin, Liu, Chan, et~al.]{tao2024maniskill3}
Stone Tao, Fanbo Xiang, Arth Shukla, Yuzhe Qin, Xander Hinrichsen, Xiaodi Yuan, Chen Bao, Xinsong Lin, Yulin Liu, Tse-kai Chan, et~al.
\newblock Maniskill3: Gpu parallelized robotics simulation and rendering for generalizable embodied ai.
\newblock \emph{arXiv preprint arXiv:2410.00425}, 2024.

\bibitem[Taylor et~al.(2022)Taylor, Kardas, Cucurull, Scialom, Hartshorn, Saravia, Poulton, Kerkez, and Stojnic]{taylor2022galactica}
Ross Taylor, Marcin Kardas, Guillem Cucurull, Thomas Scialom, Anthony Hartshorn, Elvis Saravia, Andrew Poulton, Viktor Kerkez, and Robert Stojnic.
\newblock Galactica: A large language model for science.
\newblock \emph{arXiv preprint arXiv:2211.09085}, 2022.

\bibitem[Tian et~al.(2025)Tian, Yang, Xie, Cai, Shi, Gao, Liu, Jiang, Qiu, Yuan, et~al.]{tian2025interndata}
Yang Tian, Yuyin Yang, Yiman Xie, Zetao Cai, Xu~Shi, Ning Gao, Hangxu Liu, Xuekun Jiang, Zherui Qiu, Feng Yuan, et~al.
\newblock Interndata-a1: Pioneering high-fidelity synthetic data for pre-training generalist policy.
\newblock \emph{arXiv preprint arXiv:2511.16651}, 2025.

\bibitem[Tom et~al.(2024)Tom, Schmid, Baird, Cao, Darvish, Hao, Lo, Pablo-Garc{\'\i}a, Rajaonson, Skreta, et~al.]{tom2024self}
Gary Tom, Stefan~P Schmid, Sterling~G Baird, Yang Cao, Kourosh Darvish, Han Hao, Stanley Lo, Sergio Pablo-Garc{\'\i}a, Ella~M Rajaonson, Marta Skreta, et~al.
\newblock Self-driving laboratories for chemistry and materials science.
\newblock \emph{Chemical Reviews}, 124\penalty0 (16):\penalty0 9633--9732, 2024.

\bibitem[Walke et~al.(2023)Walke, Black, Zhao, Vuong, Zheng, Hansen-Estruch, He, Myers, Kim, Du, et~al.]{walke2023bridgedata}
Homer~Rich Walke, Kevin Black, Tony~Z Zhao, Quan Vuong, Chongyi Zheng, Philippe Hansen-Estruch, Andre~Wang He, Vivek Myers, Moo~Jin Kim, Max Du, et~al.
\newblock Bridgedata v2: A dataset for robot learning at scale.
\newblock In \emph{Conference on Robot Learning}, pages 1723--1736. PMLR, 2023.

\bibitem[Wang et~al.(2023)Wang, Xian, Chen, Wang, Wang, Fragkiadaki, Erickson, Held, and Gan]{wang2023robogen}
Yufei Wang, Zhou Xian, Feng Chen, Tsun-Hsuan Wang, Yian Wang, Katerina Fragkiadaki, Zackory Erickson, David Held, and Chuang Gan.
\newblock Robogen: Towards unleashing infinite data for automated robot learning via generative simulation.
\newblock \emph{arXiv preprint arXiv:2311.01455}, 2023.

\bibitem[Wen et~al.(2025)Wen, Zhu, Li, Zhu, Tang, Wu, Xu, Liu, Cheng, Shen, et~al.]{wen2025tinyvla}
Junjie Wen, Yichen Zhu, Jinming Li, Minjie Zhu, Zhibin Tang, Kun Wu, Zhiyuan Xu, Ning Liu, Ran Cheng, Chaomin Shen, et~al.
\newblock Tinyvla: Towards fast, data-efficient vision-language-action models for robotic manipulation.
\newblock \emph{IEEE Robotics and Automation Letters}, 2025.

\bibitem[Xiang et~al.(2026)Xiang, Chen, Xu, Wang, Lv, Deng, Zhu, Dong, Zhao, Yuan, et~al.]{xiang2026native}
Jianfeng Xiang, Xiaoxue Chen, Sicheng Xu, Ruicheng Wang, Zelong Lv, Yu~Deng, Hongyuan Zhu, Yue Dong, Hao Zhao, Nicholas~Jing Yuan, et~al.
\newblock Native and compact structured latents for 3d generation.
\newblock In \emph{Proceedings of the IEEE/CVF Conference on Computer Vision and Pattern Recognition}, pages 14419--14429, 2026.

\bibitem[Yang et~al.(2025)Yang, Li, Wang, Chen, Tian, Wang, Wang, Zhao, Liao, and Pang]{yang2025instructvla}
Shuai Yang, Hao Li, Bin Wang, Yilun Chen, Yang Tian, Tai Wang, Hanqing Wang, Feng Zhao, Yiyi Liao, and Jiangmiao Pang.
\newblock Instructvla: Vision-language-action instruction tuning from understanding to manipulation.
\newblock \emph{arXiv preprint arXiv:2507.17520}, 2025.

\bibitem[Ye et~al.(2026)Ye, Ge, Zheng, Gao, Yu, Kurian, Indupuru, Tan, Zhu, Xiang, et~al.]{ye2026world}
Seonghyeon Ye, Yunhao Ge, Kaiyuan Zheng, Shenyuan Gao, Sihyun Yu, George Kurian, Suneel Indupuru, You~Liang Tan, Chuning Zhu, Jiannan Xiang, et~al.
\newblock World action models are zero-shot policies.
\newblock \emph{arXiv preprint arXiv:2602.15922}, 2026.

\bibitem[Yenamandra et~al.(2023)Yenamandra, Ramachandran, Yadav, Wang, Khanna, Gervet, Yang, Jain, Clegg, Turner, et~al.]{yenamandra2023homerobot}
Sriram Yenamandra, Arun Ramachandran, Karmesh Yadav, Austin Wang, Mukul Khanna, Theophile Gervet, Tsung-Yen Yang, Vidhi Jain, Alexander~William Clegg, John Turner, et~al.
\newblock Homerobot: Open-vocabulary mobile manipulation.
\newblock \emph{arXiv preprint arXiv:2306.11565}, 2023.

\bibitem[Yoshikawa et~al.(2022)Yoshikawa, Li, Darvish, Zhao, Xu, Kuramshin, Aspuru-Guzik, Garg, and Shkurti]{yoshikawa2022chemistry}
Naruki Yoshikawa, Andrew~Zou Li, Kourosh Darvish, Yuchi Zhao, Haoping Xu, Artur Kuramshin, Al{\'a}n Aspuru-Guzik, Animesh Garg, and Florian Shkurti.
\newblock Chemistry lab automation via constrained task and motion planning.
\newblock \emph{arXiv preprint arXiv:2212.09672}, 2022.

\bibitem[Yu et~al.(2020)Yu, Quillen, He, Julian, Hausman, Finn, and Levine]{yu2020meta}
Tianhe Yu, Deirdre Quillen, Zhanpeng He, Ryan Julian, Karol Hausman, Chelsea Finn, and Sergey Levine.
\newblock Meta-world: A benchmark and evaluation for multi-task and meta reinforcement learning.
\newblock In \emph{Conference on robot learning}, pages 1094--1100. PMLR, 2020.

\bibitem[Yue et~al.(2024)Yue, Wang, Kang, Han, Wang, Song, Feng, and Huang]{yue2024deer}
Yang Yue, Yulin Wang, Bingyi Kang, Yizeng Han, Shenzhi Wang, Shiji Song, Jiashi Feng, and Gao Huang.
\newblock Deer-vla: Dynamic inference of multimodal large language models for efficient robot execution.
\newblock \emph{Advances in Neural Information Processing Systems}, 37:\penalty0 56619--56643, 2024.

\bibitem[Zawalski et~al.(2024)Zawalski, Chen, Pertsch, Mees, Finn, and Levine]{zawalski2024robotic}
Micha{\l} Zawalski, William Chen, Karl Pertsch, Oier Mees, Chelsea Finn, and Sergey Levine.
\newblock Robotic control via embodied chain-of-thought reasoning.
\newblock \emph{arXiv preprint arXiv:2407.08693}, 2024.

\bibitem[Ze et~al.(2023)Ze, Yan, Wu, Macaluso, Ge, Ye, Hansen, Li, and Wang]{ze2023gnfactor}
Yanjie Ze, Ge~Yan, Yueh-Hua Wu, Annabella Macaluso, Yuying Ge, Jianglong Ye, Nicklas Hansen, Li~Erran Li, and Xiaolong Wang.
\newblock Gnfactor: Multi-task real robot learning with generalizable neural feature fields.
\newblock In \emph{Conference on robot learning}, pages 284--301. PMLR, 2023.

\bibitem[Zhai et~al.(2025)Zhai, Liu, Fang, Cai, Ma, Yin, Wang, Zhou, Wang, Shi, et~al.]{zhai2025igniting}
Andy Zhai, Brae Liu, Bruno Fang, Chalse Cai, Ellie Ma, Ethan Yin, Hao Wang, Hugo Zhou, James Wang, Lights Shi, et~al.
\newblock Igniting vlms toward the embodied space.
\newblock \emph{arXiv preprint arXiv:2509.11766}, 2025.

\bibitem[Zhao et~al.(2025)Zhao, Lu, Kim, Fu, Zhang, Wu, Li, Ma, Han, Finn, et~al.]{zhao2025cot}
Qingqing Zhao, Yao Lu, Moo~Jin Kim, Zipeng Fu, Zhuoyang Zhang, Yecheng Wu, Zhaoshuo Li, Qianli Ma, Song Han, Chelsea Finn, et~al.
\newblock Cot-vla: Visual chain-of-thought reasoning for vision-language-action models.
\newblock In \emph{Proceedings of the Computer Vision and Pattern Recognition Conference}, pages 1702--1713, 2025.

\bibitem[Zheng et~al.(2025{\natexlab{a}})Zheng, Li, Wang, Liu, Kang, Feng, Zheng, Zou, Chen, Zeng, et~al.]{zheng2025x}
Jinliang Zheng, Jianxiong Li, Zhihao Wang, Dongxiu Liu, Xirui Kang, Yuchun Feng, Yinan Zheng, Jiayin Zou, Yilun Chen, Jia Zeng, et~al.
\newblock X-vla: Soft-prompted transformer as scalable cross-embodiment vision-language-action model.
\newblock \emph{arXiv preprint arXiv:2510.10274}, 2025{\natexlab{a}}.

\bibitem[Zheng et~al.(2025{\natexlab{b}})Zheng, Liang, Huang, Gao, Daum{\'e}~III, Kolobov, Huang, and Yang]{zheng2025tracevla}
Ruijie Zheng, Yongyuan Liang, Shuaiyi Huang, Jianfeng Gao, Hal Daum{\'e}~III, Andrey Kolobov, Furong Huang, and Jianwei Yang.
\newblock Tracevla: Visual trace prompting enhances spatial-temporal awareness for generalist robotic policies.
\newblock In \emph{International Conference on Learning Representations}, volume 2025, pages 54277--54296, 2025{\natexlab{b}}.

\bibitem[Zhong et~al.(2025)Zhong, Yan, Li, Liu, Gong, Zhang, Song, Chen, Zheng, Wang, et~al.]{zhong2025flowvla}
Zhide Zhong, Haodong Yan, Junfeng Li, Xiangchen Liu, Xin Gong, Tianran Zhang, Wenxuan Song, Jiayi Chen, Xinhu Zheng, Hesheng Wang, et~al.
\newblock Flowvla: Visual chain of thought-based motion reasoning for vision-language-action models.
\newblock \emph{arXiv preprint arXiv:2508.18269}, 2025.

\bibitem[Zhu et~al.(2020)Zhu, Wong, Mandlekar, Mart{\'\i}n-Mart{\'\i}n, Joshi, Lin, Maddukuri, Nasiriany, and Zhu]{zhu2020robosuite}
Yuke Zhu, Josiah Wong, Ajay Mandlekar, Roberto Mart{\'\i}n-Mart{\'\i}n, Abhishek Joshi, Kevin Lin, Abhiram Maddukuri, Soroush Nasiriany, and Yifeng Zhu.
\newblock robosuite: A modular simulation framework and benchmark for robot learning.
\newblock \emph{arXiv preprint arXiv:2009.12293}, 2020.

\bibitem[Zitkovich et~al.(2023)Zitkovich, Yu, Xu, Xu, Xiao, Xia, Wu, Wohlhart, Welker, Wahid, et~al.]{zitkovich2023rt}
Brianna Zitkovich, Tianhe Yu, Sichun Xu, Peng Xu, Ted Xiao, Fei Xia, Jialin Wu, Paul Wohlhart, Stefan Welker, Ayzaan Wahid, et~al.
\newblock Rt-2: Vision-language-action models transfer web knowledge to robotic control.
\newblock In \emph{Conference on Robot Learning}, pages 2165--2183. PMLR, 2023.

\end{thebibliography}
